\documentclass{article}
\usepackage{PRIMEarxiv}
\usepackage[numbers]{natbib}
\usepackage{float}
\usepackage{microtype}
\usepackage{hyperref}
\usepackage{url}
\usepackage{lineno}
\usepackage{algorithm}
\usepackage{algorithmic}
\usepackage{graphicx}
\usepackage{subcaption}
\usepackage{amsmath,amsfonts}
\usepackage{textcomp}
\usepackage{multirow}
\usepackage{booktabs}
\usepackage{pifont}
\usepackage{array}
\usepackage{tablefootnote}
\usepackage{arydshln}
\usepackage{makecell}
\usepackage{bm}
\usepackage{caption}
\usepackage{tabularx}
\usepackage[inline]{enumitem}
\usepackage{adjustbox}
\usepackage[table]{xcolor}
\usepackage{siunitx}
\usepackage{threeparttable}
\usepackage{wrapfig}
\graphicspath{{figure/heterofusion/}}

\definecolor{oursblue}{RGB}{217,232,255}
\definecolor{groupgray}{RGB}{228,228,228}
\definecolor{avgblue}{RGB}{190,215,255}
\definecolor{darkblue}{rgb}{0,0,0.5}
\hypersetup{colorlinks=true, citecolor=darkblue, linkcolor=darkblue, urlcolor=darkblue}

\title{Can Heterogeneous Language Models Be Fused?}

\author{
  \textbf{Shilian Chen}$^{1}$, Jie Zhou$^{1,2}$\footnotemark[1], \textbf{Qin Chen}$^1$, \textbf{Wen Wu}$^1$, \textbf{Xin Li}$^2$, \textbf{Qi Feng}$^1$, \textbf{Liang He}$^1$ \\
  $^1$ School of Computer Science and Technology, East China Normal University, $^2$ Shanghai AI Laboratory \\
  \texttt{\{jzhou, qchen, lhe\}@cs.ecnu.edu.cn}, \\
  \textcolor{red}{\url{https://github.com/ECNU-ICALK/HeteroFusion}} 
}

\begin{document}

\maketitle

\begin{abstract}
Model merging aims to integrate multiple expert models into a single model that inherits their complementary strengths without incurring the inference-time cost of ensembling. Recent progress has shown that merging can be highly effective when all source models are \emph{homogeneous}, i.e., derived from the same pretrained backbone and therefore share aligned parameter coordinates or compatible task vectors. Yet this assumption is increasingly unrealistic in open model ecosystems, where useful experts are often built on different families such as Llama, Qwen, and Mistral. In such \emph{heterogeneous} settings, direct weight-space fusion becomes ill-posed due to architectural mismatch, latent basis misalignment, and amplified cross-source conflict. We address this problem with \texttt{HeteroFusion} for heterogeneous language model fusion, which consists of two key components: topology-based alignment that transfers knowledge across heterogeneous backbones by matching functional module structures instead of raw tensor coordinates, and conflict-aware denoising that suppresses incompatible or noisy transfer signals during fusion. We further provide analytical justification showing that preserving the target adapter basis while predicting structured updates leads to a stable and well-conditioned transfer process. Across heterogeneous transfer, multi-source fusion, noisy-source robustness, and cross-family generalization settings, \texttt{HeteroFusion} consistently outperforms strong merging, fusion, and ensemble baselines. 
% Our results show that heterogeneous model fusion is not only feasible, but can be made reliable when transfer is topology-aware and conflict-aware.
\end{abstract}

\section{Introduction}

Model merging aims to integrate multiple expert models into a single model that inherits their complementary strengths without incurring the runtime cost of ensembling \citep{song2026modelmergingera,li2023deepmodelfusion,yang2024modelmergingllm}. Starting from checkpoint averaging and weight-space interpolation \citep{izmailov2018averaging,garipov2018losssurfaces,draxler2018essentially,wortsman2022modelsoups}, the field has developed a broad family of techniques, including Fisher- or covariance-aware averaging \citep{matena2022fisher,jin2023dataless}, geometric and permutation-aware alignment \citep{singh2020otfusion,ainsworth2023gitrebasin,jordan2022repair,stoica2023zipit}, and task-vector-based composition \citep{ilharco2023taskarithmetic,yadav2023ties,yu2023supermario,yadav2024breadcrumbs,deep2024della,huang2024emr}. These advances show that, when expert models are sufficiently compatible, their capabilities can often be compressed into a single multitask model with little or no additional training.

However, nearly all successful merging methods operate in \emph{homogeneous} settings, where all source models are derived from the same pretrained backbone and therefore share aligned parameter coordinates, matched architectures, or at least compatible task vectors \citep{matena2022fisher,jin2023dataless,yadav2023ties,yu2023supermario}. This assumption is increasingly restrictive in open model ecosystems. In practice, useful experts are often built on different model families---for example, a Qwen expert for one task, a Llama expert for another, and a Mistral expert for a third. If model composition is to become broadly reusable, then \emph{heterogeneous} fusion, rather than same-family fusion, is the setting that matters.

Yet extending merging from homogeneous to heterogeneous experts is fundamentally challenging. Direct weight-space fusion becomes ill-posed because heterogeneous backbones do not share a common tensor topology or latent basis. Even when modules play analogous functional roles, their updates need not be expressed in compatible coordinates, and naively combining them can amplify source-specific artifacts instead of consolidating transferable knowledge. Existing heterogeneous LLM fusion approaches therefore largely bypass direct fusion through data- or logit-space transfer, such as continual distillation from multiple source models into a target model \citep{wan2024fusellm,wan2024fusechat,gao2025bohdi,yan2025infifusion,du2025skillpacks,feng2025fusionfactory}. While effective, these methods require repeated source-model querying, curated or synthetic fusion data, and additional optimization, making them closer to distillation than to direct model fusion. Inference-time ensemble methods such as GaC, PackLLM, and UniTE \citep{yu2024gac,mavromatis2024packllm,yao2024unite} avoid weight alignment altogether, but they keep all experts active at deployment and thus forfeit the simplicity and efficiency that make model merging attractive in the first place.

We argue that heterogeneous fusion is difficult for three fundamental reasons. \emph{First}, there is topology mismatch: model families differ in layer depth, hidden dimensions, attention parameterization, and MLP structure, so raw parameter arithmetic has no canonical meaning. \emph{Second}, there is latent basis misalignment: even functionally corresponding modules may represent their low-rank adaptations in different coordinate systems, making direct transfer unstable. \emph{Third}, there is cross-source conflict amplification: when multiple heterogeneous experts are fused, architecture-specific artifacts and noisy task signals can compound rather than cancel, overwhelming the signal that should be preserved. These challenges suggest that heterogeneous fusion requires more than stronger averaging rules; it needs a transfer mechanism that is both topology-aware and conflict-aware.

To this end, we propose \texttt{HeteroFusion}, a framework for heterogeneous language model fusion in adapter space. Rather than directly merging full model weights across incompatible backbones, \texttt{HeteroFusion} treats task adapters as structured carriers of transferable expertise and learns to transfer them into a target model while preserving the target adapter basis. The framework has two key components. Topology-based alignment establishes transferable correspondences across heterogeneous backbones by matching functional module structures instead of raw tensor coordinates, and predicts structured updates for the target adapter from topology-compatible source contexts. Conflict-aware denoising suppresses incompatible or noisy transfer signals before and during fusion, so that additional experts improve rather than destabilize the fused model. We further provide analytical justification showing that preserving the target adapter basis while predicting structured updates yields a stable and well-conditioned transfer process.
We evaluate \texttt{HeteroFusion} across heterogeneous transfer, multi-source fusion, noisy-source robustness, and cross-family generalization settings. Across all these scenarios, \texttt{HeteroFusion} consistently outperforms strong model-merging and heterogeneous fusion methods. 
% These results show that heterogeneous model fusion is not only feasible, but can be made reliable when transfer is topology-aware and conflict-aware.

Our contributions are threefold.
\begin{itemize}[leftmargin=*, align=left]
    \item We formulate \emph{heterogeneous language model fusion} as a distinct problem from conventional homogeneous merging, and identify its core obstacles: topology mismatch, latent basis misalignment, and cross-source conflict amplification.
    \item We propose \texttt{HeteroFusion}, a topology- and conflict-aware fusion framework that enables direct knowledge transfer and fusion across heterogeneous LLM backbones.
    \item We provide both analytical and empirical evidence for the effectiveness of \texttt{HeteroFusion}, showing strong gains over fusion and ensemble baselines across heterogeneous transfer, multi-source fusion, and cross-family generalization settings.
\end{itemize}

\section{Related Work}
\subsection{Homogeneous Model Merging}
Early work on model merging grew out of checkpoint averaging and loss-landscape geometry, showing that weight interpolation can improve generalization when solutions lie in connected low-loss regions \citep{izmailov2018averaging,garipov2018losssurfaces,draxler2018essentially}. This line later evolved into model soups \citep{wortsman2022modelsoups}, Fisher-weighted averaging \citep{matena2022fisher}, covariance-aware regression merging (RegMean) \citep{jin2023dataless}, and alignment-aware formulations based on optimal transport, permutation symmetries, and feature repair \citep{singh2020otfusion,ainsworth2023gitrebasin,jordan2022repair,stoica2023zipit}. These methods establish the core intuition that merging can work when models are sufficiently compatible in weight space, but they largely assume shared backbones or at least compatible tensor structures.

A second major thread treats fine-tuning updates as composable task vectors. Task Arithmetic \citep{ilharco2023taskarithmetic} shows that task-specific behaviors can often be added, subtracted, or scaled through simple vector operations, motivating a series of interference-aware variants such as TIES-Merging \citep{yadav2023ties}, DARE \citep{yu2023supermario}, Model Breadcrumbs \citep{yadav2024breadcrumbs}, DELLA \citep{deep2024della}, and EMR-Merging \citep{huang2024emr}. More recent work further studies adaptive or structure-aware merging, including Model Stock \citep{jang2024modelstock}, AdaMerging \citep{yang2024adamerging}, Representation Surgery \citep{yang2024representationsurgery}, Twin-Merging \citep{lu2024twinmerging}, Localize-and-Stitch \citep{he2024localizeandstitch}, Activation-Informed Merging \citep{nobari2025activationinformed}, parameter-competition balancing \citep{du2024pcb}, and evolutionary recipe search \citep{akiba2024evolutionary}. Benchmarks and surveys such as FusionBench, MergeBench, and recent surveys further systematize the space \citep{tang2024fusionbench,he2025mergebench,li2023deepmodelfusion,yang2024modelmergingllm,song2026modelmergingera}. Despite their differences, these methods overwhelmingly target \emph{homogeneous} settings: same initialization, same architecture, same tensor layout, and often same tokenizer.

% \subsection{Parameter-Efficient Fusion and Expert Composition}
% Another closely related literature studies \emph{parameter-efficient} fusion, composition, and routing. Adapters and LoRA provide modular interfaces for transferring task knowledge without modifying all backbone weights \citep{houlsby2019parameterefficient,pfeiffer2021adapterfusion,hu2022lora}. Building on this modularity, prior work explores adapter averaging or composition through AdapterSoup \citep{chronopoulou2023adaptersoup}, dynamic LoRA composition via LoraHub \citep{huang2023lorahub}, partial linearization for PEFT fusion \citep{tang2024partiallinearization}, and MoE-style or routed LoRA aggregation such as LoRAMoE and Mixture-of-LoRAs \citep{dou2024loramoe,feng2024mixtureofloras}. Related expert-composition approaches include weight-ensembling MoE \citep{tang2024wemoe}, post-hoc routing among specialists \citep{muqeeth2024phatgoose,muqeeth2023smear}, and branch-based expert training/mixing \citep{li2022branchtrainmerge,sukhbaatar2024branchtrainmix,shazeer2017outrageously,fedus2022switch}. These works show that modular parameterizations are a promising substrate for combining experts, but most of them either assume a shared backbone family or rely on routing all experts at inference time rather than exporting a single fused model.

\subsection{Heterogeneous Fusion and Model Ensemble}
The most relevant line for our setting is heterogeneous LLM fusion. Because direct weight-space merging is ill-defined across model families, recent methods primarily resort to distillation- or data-space transfer. FuseLLM \citep{wan2024fusellm} and FuseChat \citep{wan2024fusechat} distill the knowledge of structurally diverse source models into a target model through continual training and token/logit alignment. More recent heterogeneous fusion systems, such as Bohdi \citep{gao2025bohdi}, InfiFusion \citep{yan2025infifusion}, Modular SkillPacks \citep{du2025skillpacks}, and FusionFactory \citep{feng2025fusionfactory}, extend this paradigm with synthetic data generation, multi-step fusion schedules, modular capability carriers, or multi-level log-data fusion. These methods demonstrate the promise of heterogeneous capability transfer, but they are fundamentally different from direct white-box merging: they depend on source-model inference and training data pipelines, and they do not study how to make heterogeneous parameterized experts directly fuseable.

Inference-time ensembles form another alternative. Methods such as GaC, PackLLM, and UniTE aggregate outputs or token probabilities from multiple LLMs during decoding \citep{yu2024gac,mavromatis2024packllm,yao2024unite}. They are attractive because they naturally support heterogeneous tokenizers and architectures, but they retain multiple models at runtime and therefore incur extra memory and latency. In contrast, our goal is to obtain a \emph{single} fused adapter that preserves the deployment advantages of merging while still operating in a genuinely heterogeneous setting. HeteroFusion is therefore complementary to heterogeneous distillation and inference-time ensembling: it aims to make cross-architecture expert fusion possible in a white-box, adapter-space, single-model regime.

\begin{figure*}[t!]
\vspace{-1mm}
    \centering
    \scalebox{1.1}[1]{\includegraphics[width=0.90\textwidth]{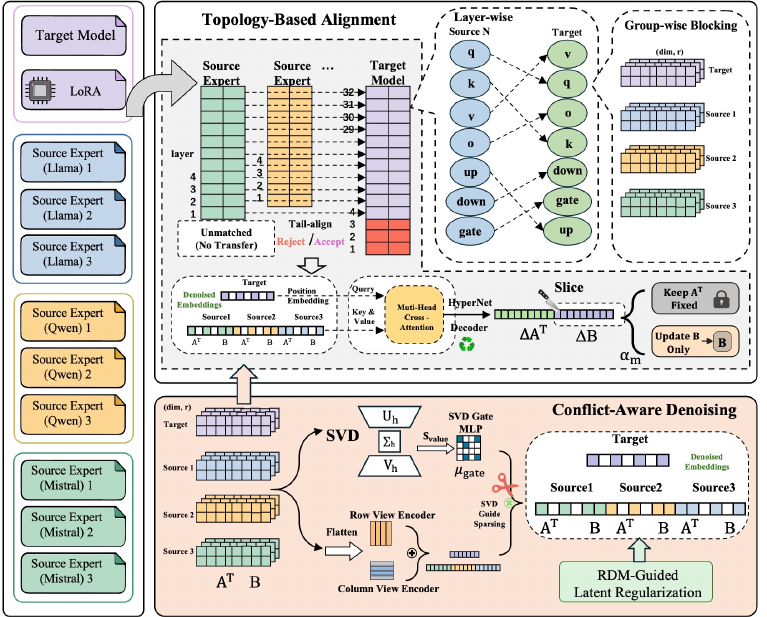}}
    % \vspace{-1mm}
    % \caption{Overview of \texttt{HeteroFusion}. The framework aligns compatible modules across heterogeneous experts, builds block-wise contexts from $A^\top$ and $B$, filters conflicting source signals, and predicts structured updates for the target $B$ matrices while preserving $A$. }
    \caption{Overview of \textbf{\texttt{HeteroFusion}}. Topology-Based Alignment maps compatible modules into unified contexts, and Conflict-Aware Denoising filters cross-source noise.}
    \label{fig:main_framework}
    % \vspace{-2.5mm}
\end{figure*}

\section{Our Methodology}
\label{sec:methodology}

Heterogeneous language model fusion is challenging because experts from different pretrained families do not share a common tensor topology or latent basis. Consequently, direct weight-space fusion is ill-posed and can corrupt the target task identity or amplify architecture-specific noise. The objective is thus to transfer reusable knowledge across heterogeneous experts while preserving the target model's task-specific structure.
To address this problem, we propose \texttt{HeteroFusion}, which comprises two components. Topology-based alignment identifies functionally compatible modules across heterogeneous backbones and specifies where transfer can occur. Conflict-aware denoising filters noisy source signals and determines what should be transferred. 
% By preserving the target adapter basis and injecting only structured updates, \texttt{HeteroFusion} achieves efficient and controlled heterogeneous fusion.

% 架构的设计逻辑(尾部对齐, 为什么只更新B ....  为什么使用行列的双特征处理, 超网络注意力 ) 稀疏化去噪的处理思路(SVD 的思路,为什么正则化,), 直接引导到附录的( Analysis of the insight of Process )
% 缺乏说明的, 为什么比各种基线和集成方法好的解释说明 

% 检查错别字, 31号晚先提交一版, 清洗记录,提交代码

\subsection{Problem Setup and Overall Framework}
\label{sec:preliminaries}

We consider a target model equipped with a target adapter $\Theta_t=\{A_t,B_t\}$ and a heterogeneous source pool $\{\Theta_s^k=\{A_s^k,B_s^k\}\}_{k=1}^{K}$, where the source adapters may come from different Transformer families and may specialize in different but related information extraction tasks. The objective is to absorb complementary knowledge from this heterogeneous pool without destroying the target model's task identity.

A key design choice in \texttt{HeteroFusion} is to preserve the target adapter basis and only predict structured corrections on the target-side $B$ matrices. Intuitively, directly modifying both low-rank factors would change the target adapter basis itself and make the transferred update harder to control. We therefore define fusion for each module group $g$ as
\begin{equation}
B_t^{g\,\prime} = B_t^g + \alpha_g \Delta B_g, 
\qquad
A_t^{g\,\prime} = A_t^g,
\label{eq:b_only_update}
\end{equation}
where $g$ indexes a group of modules with the same functional role and LoRA rank (e.g., q/k/v/o or up/down/gate), $\Delta B_g$ is a structured update predicted by the transfer network, and $\alpha_g$ is a learnable group-wise scaling factor.

To make heterogeneous adapters comparable in a unified representation space, we express each grouped LoRA update as
\begin{equation}
\Delta W_t^g = B_t^g A_t^g,
\end{equation}
and construct block-wise contexts from both $A^\top$ and $B$:
\begin{equation}
\mathcal{C}_t^g = \left[\mathrm{Blk}\!\left((A_t^g)^\top\right);\, \mathrm{Blk}(B_t^g)\right], \quad
\mathcal{C}_s^g = \mathrm{Concat}_{k=1}^{K}\!\left[\mathrm{Blk}\!\left((A_s^{k,g})^\top\right);\, \mathrm{Blk}(B_s^{k,g})\right].
\label{eq:block_context}
\end{equation}
For each block token, we further compute singular-value descriptors $s=\mathrm{svdvals}(\cdot)$ as structure-aware side information. Importantly, the source pool is concatenated rather than averaged, so the model can preserve source-specific complementarities and explicitly reason about source disagreement instead of collapsing all experts into a single mixed signal.

\subsection{Topology-based Alignment}
\label{sec:topology_alignment}

The first obstacle in heterogeneous fusion is \emph{topology mismatch}. Different model families differ in depth, hidden dimension, attention parameterization, and MLP layout, which makes raw tensor-to-tensor alignment ill-posed. For this reason, \texttt{HeteroFusion} does not attempt to align models at the level of absolute parameter coordinates. Instead, it aligns them at the level of \emph{functional topology}, where modules are matched by role and structural compatibility (see Appendix~\ref{subsec:insights_topology} for a detailed mechanistic analysis of these designs).

Specifically, for each source model $k$, we define a layer mapping $\pi_k(\ell)$ from target layer $\ell$ to a valid source layer using tail alignment and structural suffix matching. We then organize all adapter modules by \emph{module type} and \emph{rank}, which yields topology-compatible transfer units. Only module groups that admit at least one valid target--source match are activated for transfer, preventing unmatched architecture-specific parameters from entering the fusion process.

Once the valid correspondences are identified, we build target and source block contexts using Eq.~\ref{eq:block_context}. For each block token $x$, we extract complementary row and column views:
\begin{equation}
f_{\mathrm{row}}=\mathcal{P}_{\mathrm{row}}(x), \qquad
f_{\mathrm{col}}=\mathcal{P}_{\mathrm{col}}(\mathrm{TransposeView}(x)).
\end{equation}
These two views provide local structural descriptors that remain informative despite absolute shape differences. To safely fuse these features, we first filter them through our conflict-aware denoising module (detailed in Sec.~\ref{sec:conflict_denoising}) to obtain robust, denoised embeddings $\mathbf{z}_t^g$ and $\mathbf{z}_s^g$. 

The denoised target and source embeddings are then processed by a topology-aligned transfer network (HyperNet) equipped with cross-attention:
\begin{equation}
\mathbf{H}^g=\mathrm{MHA}(\mathbf{z}_t^g+\mathbf{E}_{pos},\, \mathbf{z}_s^g+\mathbf{E}_{pos},\, \mathbf{z}_s^g+\mathbf{E}_{pos}),
\end{equation}
where $\mathbf{E}_{pos}$ denotes positional encoding. The decoder outputs block-level deltas, from which only the $B$ segment is retained:
\begin{equation}
\Delta B_g=\mathrm{Slice}_B(\mathrm{Dec}(\mathbf{H}^g)),
\qquad
B_t^{g\,\prime}=B_t^g+\alpha_g\Delta B_g.
\label{eq:delta_b}
\end{equation}
In this way, topology-based alignment completes the full transfer loop: it determines \emph{where} transfer is allowed via module matching, and \emph{how} to generate structured updates via the HyperNet. Crucially, $A_t^g$ remains fixed to preserve the target adapter basis, while $\Delta B_g$ acts as a precisely routed correction that injects reusable knowledge from heterogeneous experts.

\subsection{Conflict-aware Denoising}
\label{sec:conflict_denoising}

Structural alignment alone is not sufficient. Even when modules are topologically compatible, heterogeneous experts may still provide contradictory or noisy signals due to task mismatch, architecture-specific artifacts, or source quality variation. Therefore, before the topology-aligned HyperNet performs cross-source interaction, \texttt{HeteroFusion} relies on a dedicated denoising mechanism (Appendix~\ref{subsec:insights_denoising}) to compute the robust embeddings $\mathbf{z}$.

We first apply an SVD-guided sparse gate to filter unstable channels using the singular-value descriptor $s$. Given the extracted views $f_{\mathrm{row}}$ and $f_{\mathrm{col}}$, we compute:
\begin{equation}
\mathbf{g}=\mathrm{clip}\!\left(\mathrm{ReLU}(\mathrm{MLP}(s)+\mu_{\mathrm{gate}}),\,0,\,1\right),
\quad
\mathbf{z}=\mathrm{LN}\!\left((f_{\mathrm{row}}+f_{\mathrm{col}})\odot \mathbf{g}\right).
\label{eq:svd_gate}
\end{equation}
This gate selectively suppresses channels whose spectral signatures suggest architecture-specific variation, while preserving structural directions that are more likely to be transferable. This yields the denoised target and source embeddings $\mathbf{z}_t^g$ and $\mathbf{z}_s^g$.

To further stabilize fusion under cross-source disagreement, we add a distribution-level regularizer on these hidden embeddings. We use a rectified distribution matching (RDM) loss based on a sliced Wasserstein surrogate:
\begin{equation}
\mathcal{L}_{\mathrm{rdm}}(\mathbf{z})=
\mathrm{MSE}\!\left(\mathrm{sort}(\mathbf{P}\,\mathrm{ReLU}(\mathbf{z})),
\mathrm{sort}(\mathbf{P}\,\mathrm{ReLU}(\mathbf{y}))\right),
\end{equation}
where $\mathbf{y}\sim\mathcal{N}(\mu_{\mathrm{target}},\sigma_{\mathrm{target}}^2)$ and $\mathbf{P}$ is a random projection matrix. This regularization discourages fragmented hidden distributions caused by conflicting experts, encouraging the latent space to remain well-conditioned before being fed into the HyperNet. Taken together, conflict-aware denoising ensures that only trustworthy signals survive the transfer process.

\subsection{Optimization}
\label{sec:two_stage_optimization}

Once the transfer interface is defined, we still need to optimize it without permanently overwriting the target adapter during training. To this end, we use a small amount of mixed replay data $\mathcal{D}$, which preserves complementary knowledge from source-related tasks and exposes the transfer network to the interaction between target and source expertise.

For each mini-batch, the transfer network predicts $\Delta B_g$ for every active group, the live adapter weights are \emph{temporarily} patched as $B_t^g+\alpha_g\Delta B_g$, the language modeling loss is evaluated on the patched model, and the original weights are restored immediately after the forward-backward step. This dynamic patching mechanism allows us to optimize the fusion operator based on the effect of transferred updates, without permanently corrupting the target adapter during training.

The full objective is
\begin{equation}
\min_{\phi,\{\alpha_g\}}
\mathcal{L}_{\mathrm{lm}}(\mathcal{D})
\;+\;
\lambda_{\mathrm{reg}}\,\frac{1}{G}\sum_{g=1}^{G}
\left(
\mathcal{L}_{\mathrm{rdm}}(\mathbf{z}_t^g)+
\mathcal{L}_{\mathrm{rdm}}(\mathbf{z}_s^g)
\right),
\label{eq:joint_obj}
\end{equation}
where $\phi$ denotes the parameters of the transfer network and $G$ is the number of active module groups. After training, we run the transfer network once more, keep all untouched target parameters, and export the fused adapter by preserving $A_t$ and replacing $B_t^g$ with $B_t^g+\alpha_g\Delta B_g$ on all aligned groups.

\section{Experimental Results}
\subsection{Experimental Setup} 
\paragraph{Datasets and Metrics} 
We fine-tune LLMs on the InstructUIE benchmark~\citep{wang2023instructuiemultitaskinstructiontuning} to yield specialized experts for distinct information extraction tasks (i.e., NER, RE, and ET) across various professional domains (e.g., biomedical, social media, and news). To further assess cross-family generalization, we also evaluate our framework on the GLUE benchmark~\citep{wang2018glue}. We report macro precision and macro F1-Score (F1) for the InstructUIE tasks, whereas for GLUE, we use its official task-specific metrics. 
% Further implementation details are provided in Appendix~\ref{app:implementation-details}.

\paragraph{Baselines.}
To evaluate the effectiveness of \texttt{HeteroFusion}, we compare it against three categories of baselines:
\textbf{1) Homogeneous merging baselines.} Since traditional weight-space methods cannot perform heterogeneous fusion, we evaluate them using specifically trained homogeneous experts that share the exact target backbone to provide broader comparability. We include parameter averaging (Weight Average), \textit{Task Arithmetic} \citep{ilharco2023taskarithmetic}, and interference-aware techniques like \textit{TIES-Merging} \citep{yadav2023ties}, \textit{DARE} \citep{yu2023supermario}, \textit{DELLA} \citep{deep2024della}, \textit{Breadcrumbs} \citep{yadav2024breadcrumbs}, and \textit{EMR-Merging} \citep{huang2024emr};
\textbf{2) Heterogeneous fusion/ensemble baselines.} We compare against methods that explicitly handle architectural mismatch. This includes distillation-based knowledge transfer (\textit{FuseLLM} \citep{wan2024fusellm}) and inference-time token/logit ensembling (\textit{GAC} \citep{yu2024gac}, \textit{UniTE} \citep{yao2024unite}). While these methods naturally bypass structural mismatch, they serve as heavy-weight upper bounds for runtime cost compared to our single-fused-model approach;
\textbf{3) Task-specific upper bounds.} We provide the fully supervised fine-tuning performance of both the target and source architectures to establish theoretical performance ceilings.

\paragraph{Implementation Details.}
To simulate a realistic and diverse open-model ecosystem, we construct a heterogeneous pool of expert models using the Llama 3.1 Instruct~\citep{meta2024Llama3.1}, Qwen2.5~\citep{qwen2025qwen25technicalreport}, and Mistral~\citep{jiang2023mistral7b} families. 
We train all source experts and the target anchor as LoRA adapters using Llama-Factory on NVIDIA A800-80GB GPUs. During \texttt{HeteroFusion} training, we use a lightweight mixed replay setting by sampling 300 training instances from each involved task, and train the transfer network for 3 epochs with a learning rate of $5\times10^{-5}$. We set the initial update scaling factor to $\alpha=0.3$, the gating shift to $\mu_{\mathrm{gate}}=0.10$, and the regularization weight to $\lambda_{\mathrm{reg}}=0.005$. More implementation details are provided in Appendix~\ref{app:implementation-details}.

\subsection{Main Results}

\paragraph{Single-Source Heterogeneous Transfer.}
\begin{table*}[t]
\centering
\small
\caption{Main results of single-source (Qwen to Llama) heterogeneous transfer across NER, RE, and ET benchmarks. P denotes precision. Best results are highlighted in bold.}
\label{tab:sheet_main_comparison}
% \vspace{-2mm}
\setlength{\tabcolsep}{1.2pt}
\renewcommand{\arraystretch}{1.05}
% \begin{adjustbox}{width=\textwidth}
\begin{tabular}{
l
S S
S S
S S
S S
S S
S S
S S
}
\toprule
\multirow{3}{*}{\textbf{Method}} 
& \multicolumn{4}{c}{\textbf{NER}} 
& \multicolumn{4}{c}{\textbf{RE}} 
& \multicolumn{4}{c}{\textbf{ET}} 
& \multicolumn{2}{c}{\multirow{2}{*}{\textbf{Avg}}} \\
\cmidrule(lr){2-5}\cmidrule(lr){6-9}\cmidrule(lr){10-13}
& \multicolumn{2}{c}{Mit-Movie}
& \multicolumn{2}{c}{TweetNER7}
& \multicolumn{2}{c}{New York Times}
& \multicolumn{2}{c}{CoNLL04}
& \multicolumn{2}{c}{FindVehicle}
& \multicolumn{2}{c}{FabNER}
& 
&  \\
\cmidrule(lr){2-3}\cmidrule(lr){4-5}\cmidrule(lr){6-7}\cmidrule(lr){8-9}\cmidrule(lr){10-11}\cmidrule(lr){12-13}\cmidrule(lr){14-15}
& \multicolumn{1}{c}{P} & \multicolumn{1}{c}{F1}
& \multicolumn{1}{c}{P} & \multicolumn{1}{c}{F1}
& \multicolumn{1}{c}{P} & \multicolumn{1}{c}{F1}
& \multicolumn{1}{c}{P} & \multicolumn{1}{c}{F1}
& \multicolumn{1}{c}{P} & \multicolumn{1}{c}{F1}
& \multicolumn{1}{c}{P} & \multicolumn{1}{c}{F1}
& \multicolumn{1}{c}{P} & \multicolumn{1}{c}{F1} \\
\midrule
\rowcolor{groupgray}
\multicolumn{15}{l}{\textbf{Homogeneous merging baselines}} \\
Llama Merge              & 80.85 & 79.49 & 60.47 & \bfseries 56.05 & 66.91 & 63.05 & 63.68 & 60.91 & 83.02 & 82.81 & 64.00 & 63.29 & 69.82 & 67.60 \\
EMR-Merging              & 72.85 & 55.44 & 65.48 & 42.41 & 79.92 & 71.06 & 40.04 & 35.80 & 58.58 & 57.10 & 57.63 & 53.00 & 62.42 & 52.47 \\
TIES-Merging             & 70.90 & 50.93 & \bfseries 69.57 & 43.04 & 88.47 & 81.24 & 34.34 & 31.08 & 55.66 & 48.96 & 38.08 & 29.36 & 59.50 & 47.44 \\
Breadcrumbs              & 64.96 & 48.20 & 61.62 & 37.47 & 68.88 & 60.99 & 44.73 & 40.20 & 66.63 & 66.01 & 65.94 & 63.04 & 62.13 & 52.65 \\
DELLA                    & 71.04 & 50.35 & 69.29 & 42.72 & 88.90 & \bfseries 81.82 & 31.82 & 28.78 & 48.36 & 41.41 & 36.10 & 27.81 & 57.59 & 45.48 \\
DARE (TIES+)             & 70.68 & 50.20 & 68.32 & 42.39 & \bfseries 88.92 & 81.81 & 31.31 & 28.80 & 47.67 & 40.76 & 36.08 & 27.96 & 57.16 & 45.32 \\
Task Arithmetic          & 65.55 & 47.66 & 61.21 & 36.34 & 73.61 & 65.04 & 42.78 & 37.60 & 59.22 & 58.14 & 63.70 & 60.34 & 61.01 & 50.85 \\
\midrule
\rowcolor{groupgray}
\multicolumn{15}{l}{\textbf{Heterogeneous fusion / Ensemble baselines}} \\
GAC                      & 64.62 & 60.25 & 45.34 & 42.67 & 14.56 & 15.51 & 34.74 & 34.89 & 63.99 & 56.98 & 41.12 & 36.59 & 44.06 & 41.15 \\
UniTE                    & 64.39 & 58.52 & 47.06 & 43.45 & 14.68 & 15.87 & 32.79 & 32.63 & 67.48 & 67.11 & 34.64 & 33.42 & 43.51 & 41.83 \\
FuseLLM                  & 80.88 & 78.24 & 59.97 & 55.15 & 60.93 & 52.76 & 58.26 & 55.83 & 75.32 & 74.72 & 62.50 & 62.22 & 66.31 & 63.15 \\
\rowcolor{oursblue}
% \midrule
\textbf{\texttt{HeteroFusion} (Ours)} 
                         & \bfseries 84.46 & \bfseries 83.29 & 59.24 & 55.04 & 68.46 & 62.26 & \bfseries 67.40 & \bfseries 63.48 & \bfseries 86.71 & \bfseries 86.71 & \bfseries 77.49 & \bfseries 77.48 & \bfseries 73.96 & \bfseries 71.38 \\

\midrule
\rowcolor{groupgray}
\multicolumn{15}{l}{\textbf{Task-specific upper bounds}} \\
SFT$_{\text{Llama}}$ 
                         & 85.66 & 84.72 & 65.41 & 62.56 & 92.40 & 91.00 & 71.90 & 68.96 & 79.47 & 79.47 & 83.82 & 83.83 & 79.78 & 78.42 \\

SFT$_{\text{Qwen}}$ 
                         & 86.27 & 85.17 & 65.12 & 62.40 & 90.85 & 89.50 & 75.87 & 73.34 & 98.90 & 98.90 & 97.02 & 97.02 & 85.67 & 84.39 \\
\bottomrule
\end{tabular}
% \end{adjustbox}
% \vspace{-4mm}
\end{table*}

Table~\ref{tab:sheet_main_comparison} demonstrates positive heterogeneous transfer from Qwen experts into a Llama target. First, HeteroFusion consistently improves over directly adapting the same Llama backbone with the same small transfer data, raising the average F1 from 67.60 to 71.38. This verifies that the gain does not come from additional target-side tuning, but from effective cross-architecture knowledge activation. In our framework, this advantage is mainly brought by Topology-based Alignment, which aligns compatible structures across heterogeneous backbones and converts otherwise unusable source adapters into transferable signals. Second, HeteroFusion remains clearly stronger than all merging baselines under cross-architecture mismatch, surpassing the strongest baseline, FuseLLM, by 8.23 average F1 points, while the white-box task-vector methods stay far behind. This result suggests that simple parameter arithmetic is insufficient once topology mismatch is introduced. Third, although the task-specific single-task Llama models remain the natural upper bound, HeteroFusion already recovers a substantial portion of that ceiling, showing that heterogeneous transfer can move close to dedicated single-task experts even without same-family source adapters.

\begin{table}[t]
  \centering
  \caption{Main results of multi-source heterogeneous transfer (Qwen and Mistral $\rightarrow$ Llama). }
  \label{tab:sheet_main2_mistral_setting}
  \setlength{\tabcolsep}{3.2pt}
  \renewcommand{\arraystretch}{1.05}
  \begin{adjustbox}{max width=\textwidth}
  \begin{tabular}{
  l
  S S
  S S
  S S
  S S
  S S
  S S
  S S
  }
  \toprule
  \multirow{3}{*}{\textbf{Method}} 
  & \multicolumn{4}{c}{\textbf{NER}} 
  & \multicolumn{4}{c}{\textbf{RE}} 
  & \multicolumn{4}{c}{\textbf{ET}} 
  & \multicolumn{2}{c}{\multirow{2}{*}{\textbf{Avg}}} \\
  \cmidrule(lr){2-5}\cmidrule(lr){6-9}\cmidrule(lr){10-13}
  & \multicolumn{2}{c}{MIT-Movie}
  & \multicolumn{2}{c}{TweetNER7}
  & \multicolumn{2}{c}{New York Times}
  & \multicolumn{2}{c}{CoNLL04}
  & \multicolumn{2}{c}{FindVehicle}
  & \multicolumn{2}{c}{FabNER}
  & &  \\
  \cmidrule(lr){2-3}\cmidrule(lr){4-5}\cmidrule(lr){6-7}\cmidrule(lr){8-9}\cmidrule(lr){10-11}\cmidrule(lr){12-13}\cmidrule(lr){14-15}
  & \multicolumn{1}{c}{P}  & \multicolumn{1}{c}{F1}
  & \multicolumn{1}{c}{P}  & \multicolumn{1}{c}{F1}
  & \multicolumn{1}{c}{P}  & \multicolumn{1}{c}{F1}
  & \multicolumn{1}{c}{P}  & \multicolumn{1}{c}{F1}
  & \multicolumn{1}{c}{P}  & \multicolumn{1}{c}{F1}
  & \multicolumn{1}{c}{P}  & \multicolumn{1}{c}{F1}
  & \multicolumn{1}{c}{P}  & \multicolumn{1}{c}{F1} \\
  \midrule
  
  \rowcolor{groupgray}
  \multicolumn{15}{l}{\textbf{Homogeneous merging baselines}} \\
  EMR-Merging        & 72.85 & 55.44 & 65.48 & 42.41 & 79.92 & 71.06 & 40.04 & 35.80 & 58.58 & 57.10 & 57.63 & 53.00 & 62.42 & 52.47 \\
  TIES-Merging       & 70.90 & 50.93 & \bfseries 69.57 & 43.04 & 88.47 & 81.24 & 34.34 & 31.08 & 55.66 & 48.96 & 38.08 & 29.36 & 59.50 & 47.44 \\
  Breadcrumbs        & 64.96 & 48.20 & 61.62 & 37.47 & 68.88 & 60.99 & 44.73 & 40.20 & 66.63 & 66.01 & 65.94 & 63.04 & 62.13 & 52.65 \\
  DELLA              & 71.04 & 50.35 & 69.29 & 42.72 & 88.90 & \bfseries 81.82 & 31.82 & 28.78 & 48.36 & 41.41 & 36.10 & 27.81 & 57.59 & 45.48 \\
  DARE (TIES+)       & 70.68 & 50.20 & 68.32 & 42.39 & \bfseries 88.92 & 81.81 & 31.31 & 28.80 & 47.67 & 40.76 & 36.08 & 27.96 & 57.16 & 45.32 \\
  Task Arithmetic    & 65.55 & 47.66 & 61.21 & 36.34 & 73.61 & 65.04 & 42.78 & 37.60 & 59.22 & 58.14 & 63.70 & 60.34 & 61.01 & 50.85 \\
  
  \midrule
  \rowcolor{groupgray}
  \multicolumn{15}{l}{\textbf{Heterogeneous fusion / ensemble baselines}} \\
  GAC                & 61.49 & 58.41 & 50.80 & 45.74 & 23.33 & 24.78 & 41.02 & 42.22 & 63.99 & 56.98 & 41.12 & 36.59 & 46.96 & 44.12 \\
  UniTE              & 67.53 & 61.42 & 56.68 & 48.52 & 18.33 & 18.15 & 32.25 & 30.85 & 66.72 & 65.71 & 35.08 & 28.22 & 46.10 & 42.15 \\
  FuseLLM            & 75.76 & 70.88 & 58.21 & 52.79 & 57.38 & 48.95 & 54.62 & 50.10 & 74.21 & 73.47 & 65.74 & 65.71 & 64.32 & 60.32 \\
  
  \rowcolor{oursblue}
  \textbf{\texttt{HeteroFusion} (Ours)}
                     & \bfseries 84.34 & \bfseries 83.25
                     & 62.11 & \bfseries 57.60
                     & 67.85 & 60.78
                     & \bfseries 67.89 & \bfseries 63.32
                     & \bfseries 85.39 & \bfseries 85.39
                     & \bfseries 77.51 & \bfseries 77.50
                     & \cellcolor{avgblue}\bfseries 74.18
                     & \cellcolor{avgblue}\bfseries 71.31 \\
  \bottomrule
  \end{tabular}
  \end{adjustbox}
\end{table}

\paragraph{Multi-Source Heterogeneous Transfer.}
To verify that \texttt{HeteroFusion} scales robustly beyond fortuitous expert pairings, we evaluate multi-source fusion across distinct model families simultaneously. \emph{First}, fusing Qwen and Mistral experts into a Llama target (Table~\ref{tab:sheet_main2_mistral_setting}) yields a dominant 71.31 average F1. This outperforms the strong heterogeneous baseline, FuseLLM, by 10.99 points and maintains near-lossless performance relative to the single-source setting. \emph{Second}, this target-agnostic generalization is not restricted to Llama. 
By inverting the setup to fuse Llama and Mistral experts into a Qwen target (Table~\ref{tab:sheet10_additional_comparison}, Appendix~\ref{sec:Multi-Source Heterogeneous Transfer}), \texttt{HeteroFusion} consistently secures a strong 69.84 average F1, far surpassing FuseLLM (58.66). \emph{Finally}, methodologically, this confirms our framework's synergy: Topology-based Alignment successfully bridges arbitrary structural gaps, while Conflict-aware Denoising suppresses multi-source interference, ensuring stable cross-architecture consolidation.

\subsection{Ablation Studies}
\begin{wraptable}[8]{r}{0.40\textwidth}
\vspace{-4mm}
\captionsetup{font=footnotesize,skip=4pt,justification=raggedright,singlelinecheck=false}
% \footnotesize
\centering
\caption{Average ablation results of \texttt{HeteroFusion}.}
\label{tab:sheet5_component_study_avg}
\setlength{\tabcolsep}{7pt}
\renewcommand{\arraystretch}{0.98}
\begin{tabular}{lcc}
\toprule
\textbf{Method} & \textbf{Prec} & \textbf{F1} \\
\midrule
\rowcolor{oursblue}
\textbf{\texttt{HeteroFusion}} & \cellcolor{avgblue}\textbf{74.57} & \cellcolor{avgblue}\textbf{72.20} \\
\rowcolor{groupgray}
w/o Alignment & 45.65 & 36.49 \\
\rowcolor{groupgray}
w/o Denoising & 69.81 & 66.28 \\
\bottomrule
\end{tabular}
% \vspace{-1mm}
\end{wraptable}
Table~\ref{tab:sheet5_component_study_avg} shows that both Topology-based Alignment and Conflict-aware Denoising are indispensable in \texttt{HeteroFusion}. Without Topology-based Alignment, heterogeneous fusion becomes structurally ill-posed because there is no valid cross-architecture transfer interface. We therefore report the strongest single expert among the six sources as a conservative proxy for the ``w/o Alignment'' setting. Its average F1 is only 36.49, which is dramatically lower than the 72.20 achieved by the full model. Removing Conflict-aware Denoising also causes a clear drop, reducing the average F1 to 66.28. Together, these results confirm that Topology-based Alignment is necessary to make heterogeneous transfer possible, while Conflict-aware Denoising is necessary to keep the transferred knowledge stable under cross-source conflict.

% \subsection{Ablation Studies}
% Table~\ref{tab:sheet5_component_study_avg} verifies the necessity of both components. Since removing Topology-based Alignment causes the heterogeneous fusion process to structurally fail, we report the strongest single expert among the six sources as a proxy for the ``w/o Alignment'' setting. Its average F1 of 36.49 severely lags behind \texttt{HeteroFusion} (72.20), confirming that cross-architecture fusion is fundamentally ill-posed without alignment. Meanwhile, removing Conflict-aware Denoising causes a substantial drop to 66.28 F1. This validates our design: Topology-based Alignment structurally enables knowledge transfer, whereas Conflict-aware Denoising shields it from cross-source conflict.

\subsection{Further Analysis}
\label{sec:Further Analysis}

\begin{wrapfigure}{r}{0.45\textwidth}
\vspace{-4mm}
    \centering
    \includegraphics[width=1.0\linewidth]{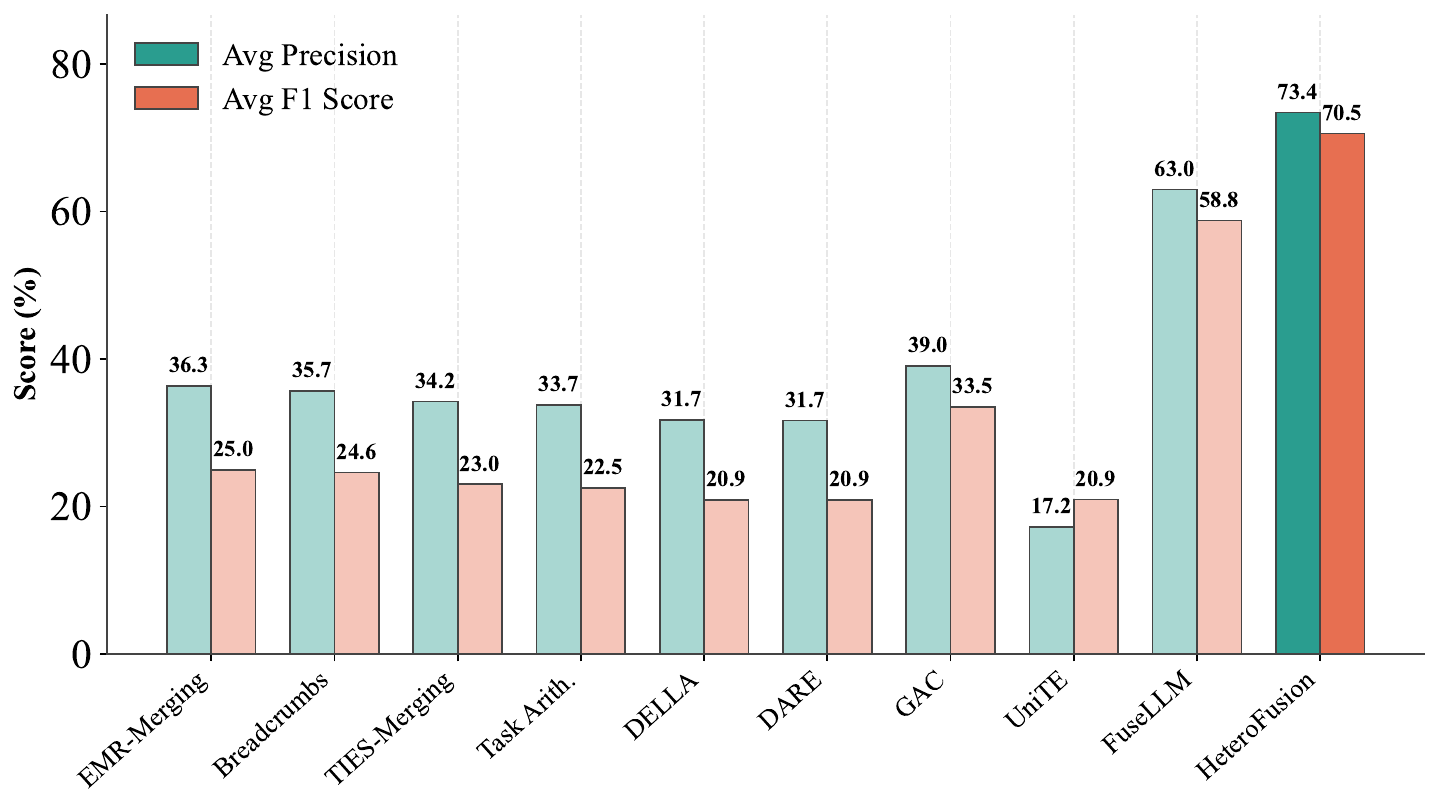}
    \vspace{-5mm}
    \caption{Results under noisy sources.}
    \label{fig:noise_setting}
    \vspace{-3mm}
\end{wrapfigure}

\paragraph{Transferability in Noisy-Source Settings}
To evaluate noise robustness, we introduce four task-irrelevant Llama experts (specializing in Chemistry, Biology, Politics, and Science) into the source pool. This shared architecture allows homogeneous merging baselines to participate but creates a highly noisy environment. As illustrated in Figure~\ref{fig:noise_setting} and detailed in Table~\ref{tab:sheet7_noise_setting}, this noise causes catastrophic interference for traditional weight-space merging methods (e.g., Task Arithmetic, DARE, TIES), plummeting their average F1 scores to $\sim$20-25\%. Inference-time ensembles similarly struggle to isolate the target task. While the distillation-based FuseLLM shows moderate resilience (58.77\% F1), \texttt{HeteroFusion} demonstrates exceptional stability, maintaining a dominant 70.55\% average F1. This robustness directly validates our conflict-aware denoising module: instead of blindly aggregating parameters, the SVD-guided sparse gate explicitly filters out conflicting spectral signatures from irrelevant experts, ensuring the target adapter only absorbs task-beneficial updates while preserving its core identity.

\begin{wrapfigure}{r}{0.45\textwidth}
% \vspace{-4mm}
    \centering
    \includegraphics[width=1.0\linewidth]{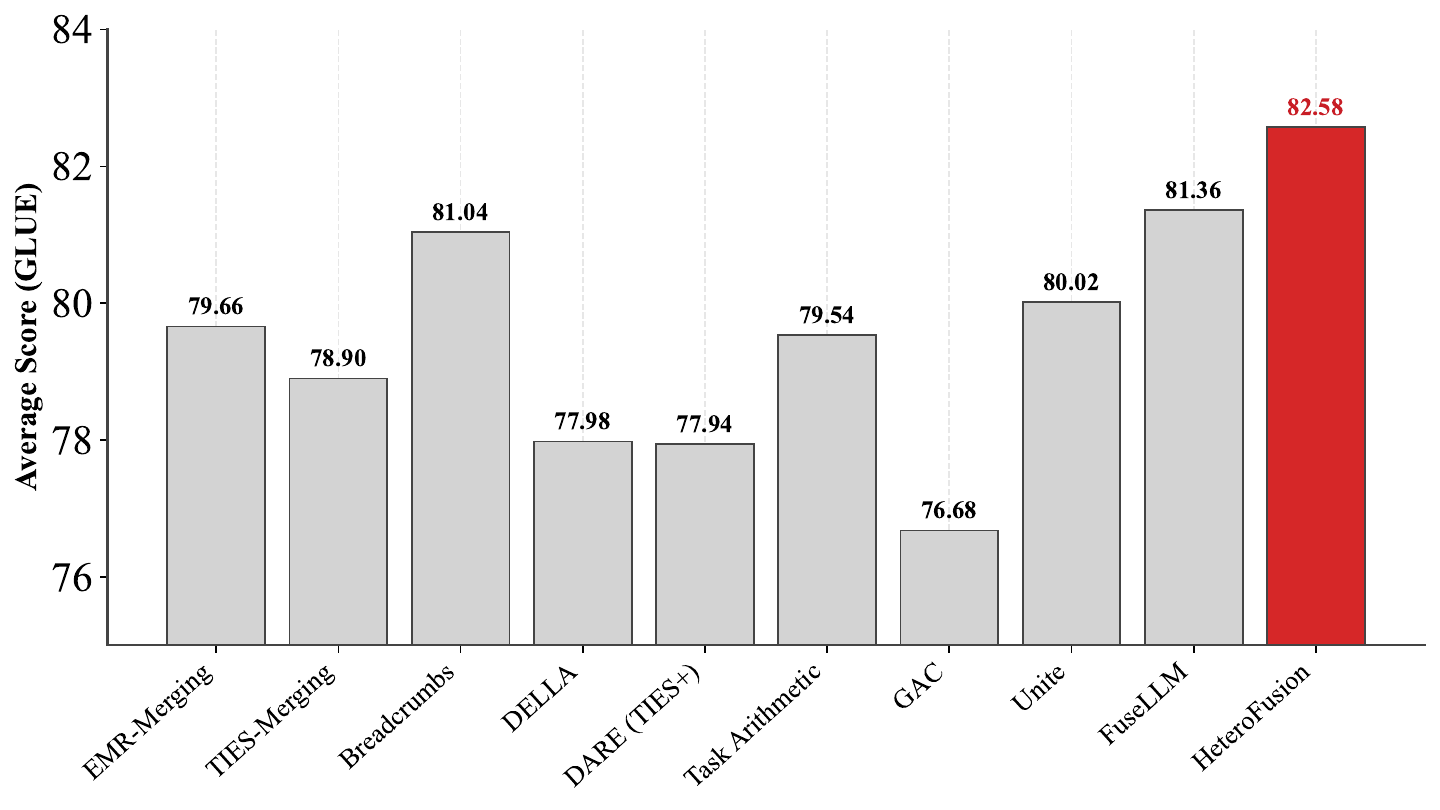}
    \vspace{-5mm}
    \caption{Average performance on the GLUE benchmark.}
    \label{fig:glue_performance}
    \vspace{-3mm}
\end{wrapfigure}

\paragraph{Transferability Across Diverse Task Families.}
To rigorously validate the broad applicability of our transfer mechanism beyond UIE-style tasks, we further evaluate \texttt{HeteroFusion} on the diverse GLUE benchmark. As illustrated in Figure~\ref{fig:glue_performance} and Table~\ref{tab:sheet8_glue_summary}, \texttt{HeteroFusion} achieves the best overall average of 82.58, outperforming FuseLLM by 1.22 points and the strongest white-box baseline, Breadcrumbs, by 1.54 points. The key point is not that our model wins every single column, but that it delivers the strongest balanced average under cross-family transfer. This confirms that the topology-aligned transfer learned by our framework is not overfitted to information extraction, while Conflict-aware Denoising successfully prevents family-specific noise from overwhelming the fused representation across entirely different task domains.

% 缺少说明以及参数选择的意义的解释,

\begin{figure}[t]
    \centering
    \includegraphics[width=0.82\textwidth]{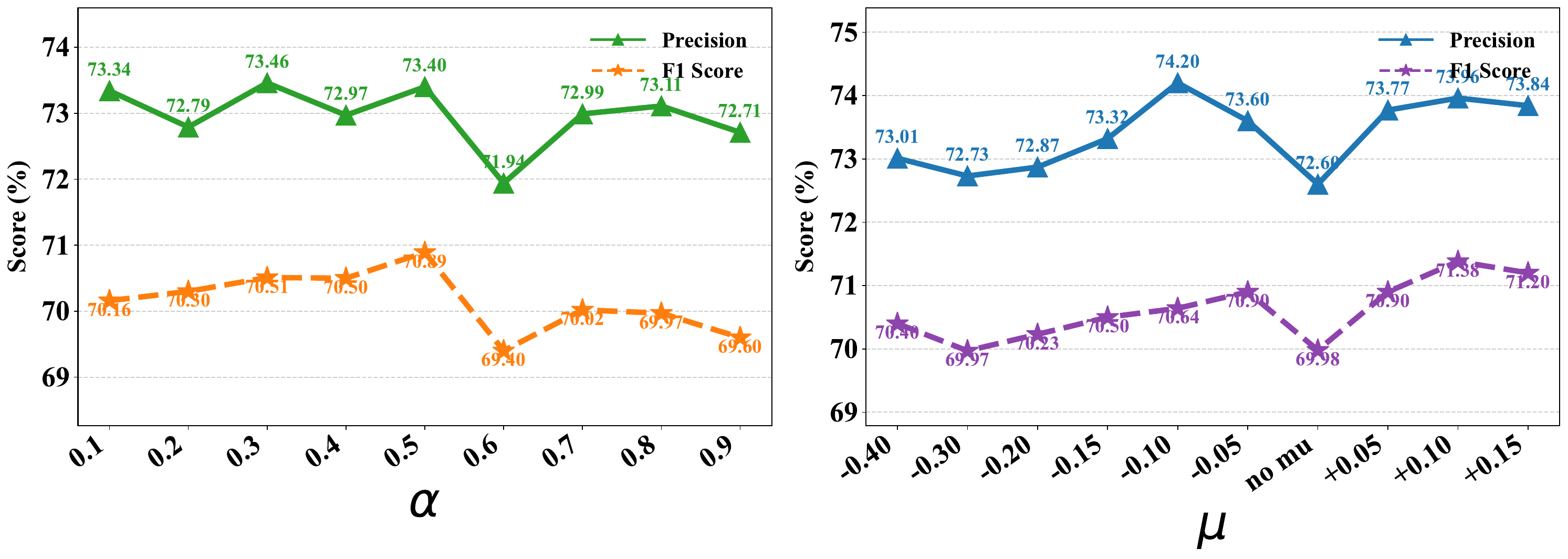}
    \caption{Sensitivity analysis of $\alpha$ and $\mu_{\mathrm{gate}}$.}
    \label{fig:sensitivity_analysis}
\end{figure}

\paragraph{Hyperparameter Sensitivity Analysis.}
As illustrated in Figure~\ref{fig:sensitivity_analysis} and further detailed in Tables~\ref{tab:sheet3_alpha_sweep} and~\ref{tab:sheet4_mu_sweep} (Appendix~\ref{app:hyperparameter-sensitivity}), \texttt{HeteroFusion} maintains a broad stable region rather than a brittle peak. Across $\alpha \in [0.05, 0.9]$, the average F1 remains tightly bounded between 69.40 and 70.89. Intuitively, low-to-mid $\alpha$ values are preferred because they inject sufficient cross-architecture knowledge without overwhelming the target task anchor. Similarly, for the gating shift $\mu_{\mathrm{gate}}$, small positive values (e.g., $0.10$ or $0.15$) yield the highest F1 scores (up to 71.38). Mechanistically, this slight positive shift moderately relaxes the denoising sparsity threshold, allowing the model to cautiously admit subtle but useful cross-family features. Unlike many merging baselines that are highly brittle to carefully tuned coefficients, our framework remains robust over a wide window. Consequently, we adopt $\alpha=0.3$ and $\mu_{\mathrm{gate}}=0.1$ as the default configuration for all our main experiments.

\begin{wrapfigure}{r}{0.4\textwidth}
\vspace{-4mm}
    \centering
    \includegraphics[width=0.4\textwidth]{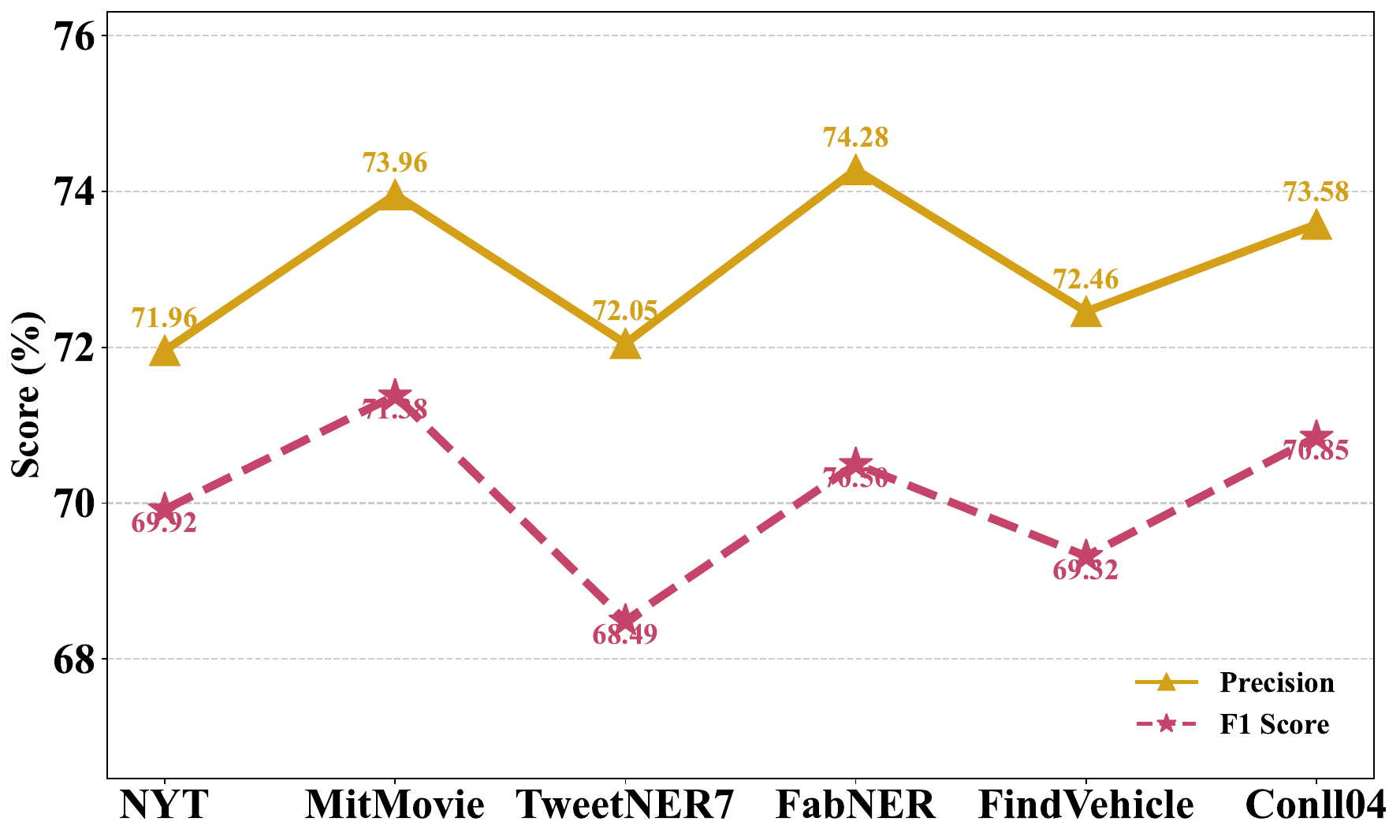}
    \vspace{-4mm}
    \caption{Influence of target anchor variants.}
    \label{fig:target_anchor_variants}
    \vspace{-4mm}
\end{wrapfigure}

\paragraph{Target Anchor Variants.}
\label{sec:target_anchor_variants}
As illustrated in Figure~\ref{fig:target_anchor_variants} and Table~\ref{tab:sheet2_base_model_study} (Appendix~\ref{app:target-anchor-variants}), we evaluate how the choice of the initial task-specific adapter impacts heterogeneous transfer.
We observe two key trends. First, while the target anchor dictates the transfer headroom—with \texttt{Mit-Movie} yielding the highest average F1 of 71.38 (our default choice)—the overall performance remains stable across all six anchors (68.49--71.38 average F1). This indicates \texttt{HeteroFusion} is robust and does not rely on a lucky initialization. Second, specific anchors consistently peak on their native datasets. This confirms that the target adapter successfully retains its core task identity as a foundation, while heterogeneous transfer strictly acts to inject complementary knowledge on top of that anchor.

\section{Conclusions and Future Work}
In this paper, we address the challenge of merging heterogeneous language models, a realistic scenario where traditional weight-space fusion fails due to structural mismatch and cross-source conflicts. We propose \texttt{HeteroFusion}, an adapter-space framework enabling direct knowledge fusion across diverse architectures like Llama, Qwen, and Mistral. By combining Topology-based Alignment with Conflict-aware Denoising, our method safely injects transferable knowledge while filtering architecture-specific noise and preserving the target's core identity. 
Extensive experiments show that \texttt{HeteroFusion} consistently outperforms strong merging, fusion, and ensemble baselines across heterogeneous transfer, multi-source fusion, noisy-source robustness, and cross-family generalization settings. 

Several promising directions remain for future research. First, although \texttt{HeteroFusion} focuses on adapter-space transfer, extending it to other parameter-efficient modules or selective full-model fusion could capture richer cross-family knowledge beyond low-rank updates. Second, the current method still relies on a small replay set to optimize the transfer operator; reducing this reliance or enabling fully data-free heterogeneous fusion would improve practicality. Third, while topology-based alignment works well across several Transformer families, more adaptive strategies may be needed for more diverse architectures, larger backbone gaps, or multimodal foundation models.
% We believe these directions will further expand the scope of heterogeneous fusion and strengthen it as a general mechanism for reusable model composition.

% There are several promising directions for future research. First, while \texttt{HeteroFusion} focuses on adapter-space transfer, it would be valuable to extend the framework to broader parameter-efficient modules or even partial full-model fusion, so as to capture richer cross-family knowledge beyond low-rank updates. Second, our current method still relies on a small replay set to optimize the transfer operator; reducing this requirement, or moving toward fully data-free heterogeneous fusion, would further improve practicality. Third, although our topology-based alignment is effective across several Transformer families, more adaptive alignment strategies may be needed for more diverse architectures, larger backbone gaps, or multimodal foundation models. 
% We believe these directions will further broaden the scope of heterogeneous fusion and strengthen its role as a general mechanism for reusable model composition.

% Extensive experiments demonstrate that \texttt{HeteroFusion} consistently outperforms existing merging and ensemble baselines across single-source, multi-source, noisy, and cross-domain settings. Future work will explore extending this framework to multimodal domains.

\bibliographystyle{unsrtnat}
\bibliography{references}

% !TEX root = main.tex
\appendix
\clearpage
\section{Theoretical and Empirical Analysis}
\label{sec:Theoretical_and_Experimental_Analysis}

We now explain why this design is stable and suitable for heterogeneous fusion.

\paragraph{Conceptual Hypothesis.}
Although different Transformer families do not share identical parameter coordinates, they still exhibit partially transferable structural regularities, such as analogous attention/MLP topologies and compatible low-rank patterns. \texttt{HeteroFusion} leverages this observation by restricting transfer to topology-compatible module groups and filtering out source-specific artifacts before they interact with the target adapter.

\paragraph{Formal Justification.}
Let the target task vector of group $g$ be $\Delta W_t^g=B_t^gA_t^g$ and the transferred one be
\begin{equation}
\Delta W_t^{g\,\prime}=(B_t^g+\alpha_g\Delta B_g)A_t^g.
\end{equation}
Then the perturbation induced by transfer satisfies
\begin{equation}
\|\Delta W_t^{g\,\prime}-\Delta W_t^g\|_F
=\|\alpha_g\Delta B_gA_t^g\|_F
\le |\alpha_g|\,\|\Delta B_g\|_F\,\|A_t^g\|_2.
\label{eq:stability_bound}
\end{equation}
Eq.~\ref{eq:stability_bound} shows that, once $A_t^g$ is fixed, the transfer magnitude is explicitly controlled by $\alpha_g$ and $\Delta B_g$. In other words, preserving the target adapter basis turns heterogeneous fusion into a constrained and interpretable perturbation process, rather than an unconstrained rewrite of the target task vector.

% % 这个写的不行
% \paragraph{Why This Proxy is Sound.}
% The core approximation in \texttt{HeteroFusion} is to use structured adapter-space transfer as a proxy for full-model heterogeneous fusion. This proxy is sound for two reasons. First, low-rank adapters localize task-specific updates into a compact and controllable subspace, which makes cross-family transfer substantially less ambiguous than directly fusing full backbone weights. Second, preserving $A_t^g$ fixes the target-side basis, so the transferred knowledge can only act as a correction within the target model's existing task geometry. Combined with topology-based alignment and conflict-aware denoising, this yields a stable fusion process in which topology determines \emph{where} transfer is allowed, basis preservation determines \emph{what} remains anchored to the target task, and denoising determines \emph{which} latent directions survive cross-source conflict. 

\section{Implementation Details}
\label{app:implementation-details}

\paragraph{Expert Model Preparation.}
Before applying our heterogeneous fusion framework, we individually train the source domain experts and the initial target anchor. All base models are fine-tuned using the open-source Llama-Factory framework\footnote{\url{https://github.com/hiyouga/Llama-Factory}} on NVIDIA A800-80GB GPUs. We fine-tune these specialized Low-Rank Adaptation (LoRA) modules, with a rank of $r=8$, for 3 epochs using a learning rate of $1\times10^{-4}$.

\paragraph{\texttt{HeteroFusion} Training Setup and Hyperparameters.}
During the fusion phase, we employ a lightweight mixed replay strategy to expose the transfer network to the interactions between target and source knowledge. Specifically, rather than using the full training corpora, we randomly sample exactly 300 instances from the training set of each involved task (including both the target anchor task and all source expert tasks) to construct the replay dataset.

The architectural and optimization hyperparameters for our framework are detailed as follows:
\begin{itemize}[leftmargin=*, align=left]
    \item \textbf{Transfer Network Architecture:} The topology-aligned cross-attention network is configured with an embedding dimension of 1024, 8 attention heads, and a maximum position embedding length of 4096 to support long-context modeling.
    \item \textbf{Training Configurations:} The transfer network is trained for 3 epochs with a learning rate of 5e-5. We set the maximum sequence cutoff length to 1024. The training utilizes a per-device batch size of 1 with 8 gradient accumulation steps. The initial group-wise scaling factor for the injected updates is set to $\alpha=0.3$.
    \item \textbf{Conflict-Aware Denoising:} For the SVD-guided sparse gate, the gating shift parameter is set to $\mu_{\mathrm{gate}}=0.10$ to cautiously admit cross-family features.
    \item \textbf{Distribution Matching Regularization:} The rectified distribution matching (RDM) regularization weight is scaled to $\lambda_{\mathrm{reg}}=0.005$ to ensure that the primary language modeling loss maintains absolute dominance during optimization. The target distribution for the sliced Wasserstein surrogate is configured with a mean of $\mu_{\mathrm{target}}=0.0$ and a standard deviation of $\sigma_{\mathrm{target}}=1.0$. The number of random projection directions used to compute the surrogate distance is set to 2048.
\end{itemize}

\section{Supplementary Experimental Tables}

\subsection{Multi-Source Heterogeneous Transfer.}
\label{sec:Multi-Source Heterogeneous Transfer}

% Table~\ref{tab:sheet10_additional_comparison} reverses the transfer direction by switching the target backbone and the expert origin. The same conclusion still holds: HeteroFusion is not tied to a particular transfer direction such as Qwen$\rightarrow$Llama. In this reciprocal setting, it achieves 69.84 average F1, outperforming FuseLLM by 11.18 points and the strongest white-box baseline by more than 13 points. More importantly, the advantage remains visible after both the alignment base and the source family are swapped, which indicates that the benefit comes from the transfer mechanism itself rather than from a specific backbone pairing. This again supports the role of Topology-based Alignment: once the structural correspondence is learned at the adapter level, the model can reuse heterogeneous experts in a direction-agnostic way.

\begin{table}[t]
  \centering
  \caption{Main results of multi-source (Llama and Mistral $\rightarrow$ Qwen) heterogeneous transfer.}
  \label{tab:sheet10_additional_comparison}
  % \vspace{-2mm}
  \setlength{\tabcolsep}{2.7pt}
  \begin{adjustbox}{max width=\textwidth}
  {\renewcommand{\arraystretch}{0.92}\setlength{\aboverulesep}{0.15ex}\setlength{\belowrulesep}{0.15ex}%
  \begin{tabular}{l cc cc cc cc cc cc c c}
    \toprule
    \multirow{3}{*}{Method} & \multicolumn{4}{c}{NER} & \multicolumn{4}{c}{RE} & \multicolumn{4}{c}{ET} & \multicolumn{2}{c}{\multirow{2}{*}{Avg}} \\
    \cmidrule(lr){2-5}\cmidrule(lr){6-9}\cmidrule(lr){10-13}
     & \multicolumn{2}{c}{Mit-Movie} & \multicolumn{2}{c}{TweetNER7} & \multicolumn{2}{c}{New York Times} & \multicolumn{2}{c}{Conll04} & \multicolumn{2}{c}{FindVehicle} & \multicolumn{2}{c}{FabNER} & & \\
    \cmidrule(lr){2-3}\cmidrule(lr){4-5}\cmidrule(lr){6-7}\cmidrule(lr){8-9}\cmidrule(lr){10-11}\cmidrule(lr){12-13}\cmidrule(lr){14-15}
     & Prec & F1 & Prec & F1 & Prec & F1 & Prec & F1 & Prec & F1 & Prec & F1 & Prec & F1 \\
    \midrule
    EMR-Merging & 78.65 & 69.88 & 65.91 & 53.58 & 58.19 & 53.12 & 45.67 & 42.09 & 68.56 & 67.21 & 48.03 & 45.05 & 60.84 & 55.16 \\
    TIES-Merging & 78.96 & 69.69 & 67.88 & 54.80 & 63.47 & 59.12 & 36.08 & 32.83 & 65.64 & 63.09 & 49.34 & 45.33 & 60.23 & 54.14 \\
    Breadcrumbs & 73.35 & 67.69 & 54.07 & 47.33 & 22.97 & 22.17 & 45.96 & 43.62 & 66.84 & 66.26 & 42.72 & 41.42 & 50.99 & 48.08 \\
    DELLA & 79.47 & 66.72 & \textbf{71.77} & \textbf{55.62} & 73.43 & 69.57 & 29.80 & 26.60 & 70.42 & 65.21 & 59.71 & 52.93 & 64.10 & 56.11 \\
    DARE (TIES+) & 78.70 & 66.30 & 71.58 & 55.58 & \textbf{73.98} & \textbf{69.95} & 30.34 & 27.28 & 70.41 & 64.41 & 59.25 & 52.70 & 64.04 & 56.04 \\
    Task Arithmetic & 74.82 & 68.98 & 57.03 & 49.72 & 27.92 & 26.45 & 48.30 & 45.61 & 67.39 & 66.76 & 43.74 & 42.14 & 53.20 & 49.94 \\
    \midrule
    UniTE & 67.53 & 61.42 & 56.68 & 48.52 & 18.33 & 18.15 & 32.25 & 30.85 & 66.72 & 65.71 & 35.08 & 28.22 & 46.10 & 42.15 \\
    FuseLLM & 79.25 & 75.08 & 60.30 & 50.76 & 47.12 & 38.87 & 62.91 & 57.47 & 73.19 & 73.09 & 57.00 & 56.70 & 63.30 & 58.66 \\
    \textbf{HeteroFusion} & \textbf{83.58} & \textbf{82.62} & 57.71 & 54.38 & 63.87 & 57.75 & \textbf{67.40} & \textbf{63.95} & \textbf{84.47} & \textbf{84.47} & \textbf{75.83} & \textbf{75.84} & \textbf{72.14} & \textbf{69.84} \\
    \bottomrule
  \end{tabular}}
  \end{adjustbox}
\end{table}

\begin{table}[t]
  \centering
  \caption{Detailed configuration of target anchors and source experts for multi-source heterogeneous transfer experiments.}
  \label{tab:expert_configuration}
  \renewcommand{\arraystretch}{1.1}
  % \begin{adjustbox}{max width=\textwidth}
  \begin{tabular}{lllc}
  \toprule
  \textbf{Experiment Setup} & \textbf{Role} & \textbf{Base Model} & \textbf{Task} \\
  \midrule
  \multirow{6}{*}{\makecell[l]{\textbf{Llama Target Fusion}\\(Corresponding to Table \ref{tab:sheet_main2_mistral_setting})}} 
  & Target & Llama-3.1-8B-Instruct &  mit-movie \\
  & Source & Qwen2.5-7B-Instruct   &  TweetNER7 \\
  & Source & Qwen2.5-7B-Instruct   &  conll04 \\
  & Source & Qwen2.5-7B-Instruct   &  New York Times \\
  & Source & Mistral-7b-v0.3       &  FabNER \\
  & Source & Mistral-7b-v0.3       &  FindVehicle \\
  \midrule
  \multirow{6}{*}{\makecell[l]{\textbf{Qwen Target Fusion}\\(Corresponding to Table \ref{tab:sheet10_additional_comparison})}} 
  & Target & Qwen2.5-7B-Instruct   &  mit-movie \\
  & Source & Llama-3.1-8B-Instruct &  TweetNER7 \\
  & Source & Llama-3.1-8B-Instruct &  conll04 \\
  & Source & Llama-3.1-8B-Instruct &  New York Times \\
  & Source & Mistral-7b-v0.3       &  FabNER \\
  & Source & Mistral-7b-v0.3       &  FindVehicle \\
  \bottomrule
  \end{tabular}
  % \end{adjustbox}
\end{table}

% =========================================================================
% 核心修复点：强制输出上面的表格！不排完绝对不准进入 C.2！
% 这会在表格排不下时插入分页，但保证 C.2 绝对在所有表格的下方开始。
% =========================================================================
\clearpage

\subsection{Details of the Ablation Study}
\begin{table}[H]
\centering
\caption{Ablation study of \texttt{HeteroFusion}. Removing either topology-based alignment or conflict-aware denoising leads to a clear drop in average performance. Best results are shown in \textbf{bold}.}
\label{tab:sheet5_component_study}
\begin{threeparttable}
\setlength{\tabcolsep}{3.0pt}
\renewcommand{\arraystretch}{1.05}

\begin{adjustbox}{width=\textwidth}
\begin{tabular}{
l
S S
S S
S S
S S
S S
S S
S S
}
\toprule
\multirow{3}{*}{\textbf{Method}} 
& \multicolumn{4}{c}{\textbf{NER}} 
& \multicolumn{4}{c}{\textbf{RE}} 
& \multicolumn{4}{c}{\textbf{ET}} 
& \multicolumn{2}{c}{\multirow{2}{*}{Avg}}  \\
\cmidrule(lr){2-5}\cmidrule(lr){6-9}\cmidrule(lr){10-13}
& \multicolumn{2}{c}{MIT-Movie}
& \multicolumn{2}{c}{TweetNER7}
& \multicolumn{2}{c}{New York Times}
& \multicolumn{2}{c}{CoNLL04}
& \multicolumn{2}{c}{FindVehicle}
& \multicolumn{2}{c}{FabNER}
& &  \\
\cmidrule(lr){2-3}\cmidrule(lr){4-5}\cmidrule(lr){6-7}\cmidrule(lr){8-9}\cmidrule(lr){10-11}\cmidrule(lr){12-13}\cmidrule(lr){14-15}
& \multicolumn{1}{c}{P}  & \multicolumn{1}{c}{F1}
& \multicolumn{1}{c}{P}  & \multicolumn{1}{c}{F1}
& \multicolumn{1}{c}{P}  & \multicolumn{1}{c}{F1}
& \multicolumn{1}{c}{P}  & \multicolumn{1}{c}{F1}
& \multicolumn{1}{c}{P}  & \multicolumn{1}{c}{F1}
& \multicolumn{1}{c}{P}  & \multicolumn{1}{c}{F1}
& \multicolumn{1}{c}{P}  & \multicolumn{1}{c}{F1} \\
\midrule
\rowcolor{oursblue}
\textbf{\texttt{HeteroFusion}}
& 84.92 & 83.81
& \bfseries 61.46 & \bfseries 56.45
& \bfseries 70.32 & \bfseries 64.80
& 65.55 & 63.01
& \bfseries 87.26 & \bfseries 87.26
& \bfseries 77.89 & \bfseries 77.87
& \cellcolor{avgblue}\bfseries 74.57
& \cellcolor{avgblue}\bfseries 72.20 \\

\rowcolor{groupgray}
\makecell[l]{w/o Alignment}
& 50.19 & 36.42
& 40.66 & 27.68
& 9.17 & 7.99
& \bfseries 72.09 & \bfseries 69.33
& 59.77 & 48.19
& 42.00 & 29.35
& 45.65 & 36.49 \\

\rowcolor{groupgray}
\makecell[l]{w/o Denoising}
& \bfseries 85.80 & \bfseries 84.49
& 60.37 & 54.07
& 64.74 & 55.42
& 53.03 & 48.96
& 84.84 & 84.70
& 70.09 & 70.03
& 69.81 & 66.28 \\
\bottomrule
\end{tabular}
\end{adjustbox}
\end{threeparttable}

% \vspace{-3mm}
\end{table}

\subsection{Hyperparameter Sensitivity Analysis}
\label{app:hyperparameter-sensitivity}

\begin{table}[H]
  \caption{Alpha sweep.}
  \label{tab:sheet3_alpha_sweep}
  % \vspace{-2mm}
  \centering
  \setlength{\tabcolsep}{2.7pt}
  \resizebox{\textwidth}{!}{
  {\renewcommand{\arraystretch}{0.92}\setlength{\aboverulesep}{0.15ex}\setlength{\belowrulesep}{0.15ex}%
  \begin{tabular}{l cc cc cc cc cc cc c c}
    \toprule
    \multirow{3}{*}{Alpha} & \multicolumn{4}{c}{NER} & \multicolumn{4}{c}{RE} & \multicolumn{4}{c}{ET} & \multicolumn{2}{c}{\multirow{2}{*}{Avg}} \\
    \cmidrule(lr){2-5}\cmidrule(lr){6-9}\cmidrule(lr){10-13}
     & \multicolumn{2}{c}{Mit-Movie} & \multicolumn{2}{c}{TweetNER7} & \multicolumn{2}{c}{New York Times} & \multicolumn{2}{c}{Conll04} & \multicolumn{2}{c}{FindVehicle} & \multicolumn{2}{c}{FabNER} & & \\
    \cmidrule(lr){2-3}\cmidrule(lr){4-5}\cmidrule(lr){6-7}\cmidrule(lr){8-9}\cmidrule(lr){10-11}\cmidrule(lr){12-13}\cmidrule(lr){14-15}
     & Prec & F1 & Prec & F1 & Prec & F1 & Prec & F1 & Prec & F1 & Prec & F1 & Prec & F1 \\
    \midrule
    \texttt{0p05} & \textbf{84.96} & \textbf{83.86} & 59.16 & 55.53 & 68.41 & \textbf{62.64} & 62.55 & 60.12 & 86.40 & 86.38 & 75.59 & 75.60 & 72.85 & 70.69 \\
    \texttt{0p1} & 84.19 & 83.02 & 62.27 & 55.61 & 67.53 & 60.12 & 64.30 & 60.48 & 85.68 & 85.66 & 76.04 & 76.04 & 73.34 & 70.16 \\
    \texttt{0p15} & 84.65 & 83.61 & 58.33 & 54.70 & 66.31 & 59.67 & 66.41 & \textbf{62.64} & 83.93 & 83.93 & 76.01 & 76.03 & 72.61 & 70.10 \\
    \texttt{0p2} & 83.98 & 82.90 & 59.96 & 56.18 & 66.28 & 59.58 & 63.43 & 60.08 & \textbf{86.48} & \textbf{86.48} & \textbf{76.58} & \textbf{76.58} & 72.79 & 70.30 \\
    \texttt{0p25} & 83.90 & 82.79 & 61.19 & 55.87 & 68.83 & 62.14 & 61.80 & 58.21 & 85.78 & 85.77 & 75.03 & 75.05 & 72.76 & 69.97 \\
    \texttt{0p3} & 83.91 & 82.85 & 61.38 & 56.14 & \textbf{69.04} & 61.47 & 64.18 & 60.34 & 86.45 & 86.44 & 75.77 & 75.79 & \textbf{73.46} & 70.51 \\
    \texttt{0p35} & 84.20 & 83.19 & 58.68 & 55.21 & 68.36 & 62.25 & 63.08 & 59.56 & 86.13 & 86.13 & 75.92 & 75.91 & 72.73 & 70.38 \\
    \texttt{0p4} & 83.92 & 82.70 & 61.36 & 56.46 & 67.64 & 62.43 & 63.35 & 59.88 & 85.44 & 85.44 & 76.08 & 76.07 & 72.97 & 70.50 \\
    \texttt{0p45} & 84.12 & 83.00 & 62.51 & 56.22 & 68.52 & 60.33 & 62.55 & 57.74 & 86.19 & 86.18 & 75.27 & 75.28 & 73.19 & 69.79 \\
    \texttt{0p5} & 83.04 & 82.20 & 59.97 & 55.98 & 68.52 & 62.19 & 67.22 & \textbf{63.34} & 85.58 & 85.58 & 76.04 & 76.04 & 73.40 & \textbf{70.89} \\
    \texttt{0p55} & 83.28 & 82.23 & 59.47 & 54.69 & 66.74 & 59.37 & 65.95 & 61.98 & 85.90 & 85.91 & 74.71 & 74.72 & 72.68 & 69.82 \\
    \texttt{0p6} & 84.46 & 83.15 & 57.89 & 54.65 & 67.48 & 60.01 & 62.01 & 58.82 & 85.36 & 85.36 & 74.43 & 74.40 & 71.94 & 69.40 \\
    \texttt{0p65} & 83.25 & 82.07 & 61.32 & 56.45 & 68.02 & 59.97 & 63.72 & 60.09 & 84.84 & 84.84 & 75.40 & 75.42 & 72.76 & 69.81 \\
    \texttt{0p7} & 83.22 & 82.16 & 61.43 & 56.22 & 67.66 & 59.55 & 64.39 & 60.98 & 85.62 & 85.63 & 75.60 & 75.58 & 72.99 & 70.02 \\
    \texttt{0p75} & 83.63 & 82.33 & 59.24 & 55.61 & 67.62 & 60.71 & 65.48 & 61.31 & 85.51 & 85.51 & 76.04 & 76.06 & 72.92 & 70.26 \\
    \texttt{0p8} & 84.63 & 83.24 & \textbf{63.57} & \textbf{56.68} & 65.81 & 59.01 & 63.76 & 60.01 & 84.85 & 84.84 & 76.06 & 76.06 & 73.11 & 69.97 \\
    \texttt{0p85} & 83.51 & 82.53 & 57.58 & 53.93 & 67.05 & 59.49 & \textbf{68.15} & 63.20 & 85.60 & 85.60 & 75.62 & 75.61 & 72.92 & 70.06 \\
    \texttt{0p9} & 83.23 & 81.78 & 59.16 & 53.86 & 68.26 & 60.39 & 64.18 & 60.14 & 86.17 & 86.17 & 75.26 & 75.25 & 72.71 & 69.60 \\
    \bottomrule
  \end{tabular}}
  }
  % \vspace{-1mm}
\end{table}

\begin{table}[H]
  \caption{Mu sweep.}
  \label{tab:sheet4_mu_sweep}
  % \vspace{-2mm}
  \centering
  \setlength{\tabcolsep}{2.5pt}
  \resizebox{\textwidth}{!}{
  {\renewcommand{\arraystretch}{0.92}\setlength{\aboverulesep}{0.15ex}\setlength{\belowrulesep}{0.15ex}%
  \begin{tabular}{l cc cc cc cc cc cc c c}
    \toprule
    \multirow{3}{*}{Setting} & \multicolumn{4}{c}{NER} & \multicolumn{4}{c}{RE} & \multicolumn{4}{c}{ET} & \multicolumn{2}{c}{\multirow{2}{*}{Avg}} \\
    \cmidrule(lr){2-5}\cmidrule(lr){6-9}\cmidrule(lr){10-13}
     & \multicolumn{2}{c}{Mit-Movie} & \multicolumn{2}{c}{TweetNER7} & \multicolumn{2}{c}{New York Times} & \multicolumn{2}{c}{Conll04} & \multicolumn{2}{c}{FindVehicle} & \multicolumn{2}{c}{FabNER} & & \\
    \cmidrule(lr){2-3}\cmidrule(lr){4-5}\cmidrule(lr){6-7}\cmidrule(lr){8-9}\cmidrule(lr){10-11}\cmidrule(lr){12-13}\cmidrule(lr){14-15}
     & Prec & F1 & Prec & F1 & Prec & F1 & Prec & F1 & Prec & F1 & Prec & F1 & Prec & F1 \\
    \midrule
    \texttt{alpha\_0p3\_mu\_m0p05} & 83.44 & 82.52 & 61.57 & 56.69 & 68.79 & 62.54 & 66.93 & 62.81 & 85.05 & 85.03 & 75.85 & 75.84 & 73.60 & 70.90 \\
    \texttt{alpha\_0p3\_mu\_m0p10} & 84.25 & 83.03 & 61.61 & 54.65 & \textbf{70.90} & \textbf{62.84} & 66.88 & 61.70 & 85.69 & 85.69 & 75.89 & 75.90 & \textbf{74.20} & 70.64 \\
    \texttt{alpha\_0p3\_mu\_m0p15} & 83.53 & 82.41 & 59.14 & 54.46 & 67.03 & 60.91 & \textbf{67.57} & 62.56 & 86.56 & 86.56 & 76.11 & 76.10 & 73.32 & 70.50 \\
    \texttt{alpha\_0p3\_mu\_m0p20} & 83.91 & 82.78 & 59.44 & 54.15 & 68.21 & 61.80 & 64.79 & 61.80 & 85.36 & 85.36 & 75.50 & 75.51 & 72.87 & 70.23 \\
    \texttt{alpha\_0p3\_mu\_m0p30} & 84.66 & 83.43 & 59.36 & 54.52 & 65.91 & 58.99 & 66.05 & 62.42 & 85.63 & 85.63 & 74.78 & 74.82 & 72.73 & 69.97 \\
    \texttt{alpha\_0p3\_mu\_m0p40} & 83.82 & 82.81 & 58.52 & 55.14 & 67.91 & 59.98 & 67.00 & \textbf{63.68} & 86.12 & 86.12 & 74.69 & 74.70 & 73.01 & 70.40 \\
    \texttt{alpha\_0p3\_mu\_p0p05} & \textbf{84.84} & 83.61 & 61.05 & 55.17 & 68.30 & 61.92 & 65.88 & 62.14 & 85.86 & 85.86 & 76.69 & 76.70 & 73.77 & 70.90 \\
    \texttt{alpha\_0p3\_mu\_p0p10} & 84.46 & 83.29 & 59.24 & 55.04 & 68.46 & 62.26 & 67.40 & 63.48 & \textbf{86.71} & \textbf{86.71} & 77.49 & 77.48 & 73.96 & \textbf{71.38} \\
    \texttt{alpha\_0p3\_mu\_p0p15} & 84.82 & \textbf{83.81} & \textbf{61.99} & \textbf{56.77} & 68.32 & 62.41 & 64.46 & 60.80 & 85.62 & 85.62 & \textbf{77.82} & \textbf{77.81} & 73.84 & 71.20 \\
    \bottomrule
  \end{tabular}}
  }
  % \vspace{-1mm}
\end{table}

\subsection{Target Anchor Variants}
\label{app:target-anchor-variants}

\begin{table}[H]
  \caption{Target-anchor study under different base choices.}
  \label{tab:sheet2_base_model_study}
  % \vspace{-2mm}
  \centering
  \setlength{\tabcolsep}{3.0pt}
  \resizebox{\textwidth}{!}{
  {\renewcommand{\arraystretch}{0.92}\setlength{\aboverulesep}{0.15ex}\setlength{\belowrulesep}{0.15ex}%
  \begin{tabular}{l cc cc cc cc cc cc c c}
    \toprule
    \multirow{3}{*}{Base Model} & \multicolumn{4}{c}{NER} & \multicolumn{4}{c}{RE} & \multicolumn{4}{c}{ET} & \multicolumn{2}{c}{\multirow{2}{*}{Avg}} \\
    \cmidrule(lr){2-5}\cmidrule(lr){6-9}\cmidrule(lr){10-13}
     & \multicolumn{2}{c}{Mit-Movie} & \multicolumn{2}{c}{TweetNER7} & \multicolumn{2}{c}{New York Times} & \multicolumn{2}{c}{Conll04} & \multicolumn{2}{c}{FindVehicle} & \multicolumn{2}{c}{FabNER} & & \\
    \cmidrule(lr){2-3}\cmidrule(lr){4-5}\cmidrule(lr){6-7}\cmidrule(lr){8-9}\cmidrule(lr){10-11}\cmidrule(lr){12-13}\cmidrule(lr){14-15}
     & Prec & F1 & Prec & F1 & Prec & F1 & Prec & F1 & Prec & F1 & Prec & F1 & Prec & F1 \\
    \midrule
    \texttt{BaseMitMovie} & \textbf{84.46} & \textbf{83.29} & 59.24 & 55.04 & 68.46 & 62.26 & 67.40 & 63.48 & \textbf{86.71} & \textbf{86.71} & 77.49 & 77.48 & 73.96 & \textbf{71.38} \\
    \texttt{BaseConll04} & 81.57 & 79.61 & 60.90 & 56.63 & 68.14 & 60.56 & \textbf{71.46} & \textbf{68.93} & 86.09 & 86.08 & 73.31 & 73.31 & 73.58 & 70.85 \\
    \texttt{BaseFabNER} & 80.70 & 78.61 & 62.35 & 55.13 & 67.96 & 58.48 & 65.91 & 62.05 & 85.69 & 85.69 & \textbf{83.04} & \textbf{83.02} & \textbf{74.28} & 70.50 \\
    \texttt{BaseFindVehicle} & 79.90 & 78.26 & 60.69 & 54.53 & 66.94 & 59.52 & 66.82 & 63.18 & 84.94 & 84.94 & 75.46 & 75.46 & 72.46 & 69.32 \\
    \texttt{BaseNewYorkTimesRE} & 79.35 & 78.45 & 55.85 & 53.47 & \textbf{69.22} & \textbf{63.68} & 63.58 & 60.13 & 85.92 & 85.92 & 77.84 & 77.85 & 71.96 & 69.92 \\
    \texttt{BaseTweetNER7} & 81.25 & 78.97 & \textbf{62.91} & \textbf{56.83} & 66.05 & 57.20 & 62.81 & 58.67 & 84.90 & 84.90 & 74.38 & 74.37 & 72.05 & 68.49 \\
    \bottomrule
  \end{tabular}}
  }
  \vspace{-1mm}
\end{table}

\subsection{Noise Robustness}
\begin{table}[H]
  \caption{Results under the noisy-source setting.}
  \label{tab:sheet7_noise_setting}
  \vspace{-2mm}
  \centering
  \setlength{\tabcolsep}{2.7pt}
  \resizebox{\textwidth}{!}{
  {\renewcommand{\arraystretch}{0.92}\setlength{\aboverulesep}{0.15ex}\setlength{\belowrulesep}{0.15ex}%
  \begin{tabular}{l cc cc cc cc cc cc c c}
    \toprule
    \multirow{3}{*}{Method} & \multicolumn{4}{c}{NER} & \multicolumn{4}{c}{RE} & \multicolumn{4}{c}{ET} & \multicolumn{2}{c}{\multirow{2}{*}{Avg}} \\
    \cmidrule(lr){2-5}\cmidrule(lr){6-9}\cmidrule(lr){10-13}
     & \multicolumn{2}{c}{Mit-Movie} & \multicolumn{2}{c}{TweetNER7} & \multicolumn{2}{c}{New York Times} & \multicolumn{2}{c}{Conll04} & \multicolumn{2}{c}{FindVehicle} & \multicolumn{2}{c}{FabNER} & & \\
    \cmidrule(lr){2-3}\cmidrule(lr){4-5}\cmidrule(lr){6-7}\cmidrule(lr){8-9}\cmidrule(lr){10-11}\cmidrule(lr){12-13}\cmidrule(lr){14-15}
     & Prec & F1 & Prec & F1 & Prec & F1 & Prec & F1 & Prec & F1 & Prec & F1 & Prec & F1 \\
    \midrule
    EMR-Merging & 68.69 & 47.39 & 54.89 & 44.36 & 10.11 & 9.17 & 0.87 & 0.87 & 48.80 & 28.14 & 34.68 & 19.90 & 36.34 & 24.97 \\
    Breadcrumbs & 57.88 & 40.00 & 51.12 & 39.63 & 3.45 & 3.06 & 2.16 & 2.16 & 58.56 & 37.27 & 40.83 & 25.54 & 35.67 & 24.61 \\
    TIES-Merging & 60.95 & 41.34 & 52.53 & 40.08 & 7.97 & 7.29 & 0.87 & 0.87 & 47.94 & 28.16 & 35.24 & 20.32 & 34.25 & 23.01 \\
    Task Arithmetic & 58.52 & 39.93 & 50.22 & 38.41 & 2.34 & 1.98 & 2.16 & 2.16 & 52.91 & 31.10 & 36.30 & 21.46 & 33.74 & 22.51 \\
    DELLA & 59.93 & 40.45 & 50.97 & 38.45 & 4.88 & 4.47 & 0.43 & 0.43 & 42.55 & 23.45 & 31.47 & 17.82 & 31.71 & 20.85 \\
    DARE (TIES+) & 59.54 & 40.05 & 50.94 & 38.27 & 5.17 & 4.73 & 0.87 & 0.87 & 41.93 & 23.56 & 31.55 & 17.74 & 31.67 & 20.87 \\
    \midrule
    GAC & 59.33 & 54.11 & 42.86 & 38.38 & 8.79 & 8.43 & 21.21 & 18.48 & 66.15 & 51.11 & 35.83 & 30.64 & 39.03 & 33.53 \\
    UniTE & 24.87 & 32.22 & 18.03 & 23.83 & 2.94 & 3.78 & 11.86 & 12.90 & 35.50 & 40.77 & 10.09 & 12.22 & 17.22 & 20.95 \\
    FuseLLM & 75.74 & 70.97 & 57.17 & 51.54 & 52.10 & 43.27 & 54.87 & 50.19 & 75.67 & 74.74 & 62.21 & 61.92 & 62.96 & 58.77 \\
    \textbf{HeteroFusion} & \textbf{85.05} & \textbf{83.65} & \textbf{62.08} & \textbf{56.73} & \textbf{68.86} & \textbf{61.87} & \textbf{61.51} & \textbf{58.01} & \textbf{85.92} & \textbf{85.90} & \textbf{77.09} & \textbf{77.11} & \textbf{73.42} & \textbf{70.55} \\
    \bottomrule
  \end{tabular}}
  }
  % \vspace{-1mm}
\end{table}

\begin{table}[H]
  \centering
  \caption{Detailed configuration of target anchor and source experts for the noise robustness experiment. Four task-irrelevant, same-family (Llama) experts are introduced as noise to evaluate conflict-aware denoising.}
  \label{tab:noise_expert_configuration}
  \renewcommand{\arraystretch}{1.1}
  \begin{tabular}{llll}
    \toprule
    \textbf{Role} & \textbf{Base Model} & \textbf{Task} & \textbf{Type} \\
    \midrule
    Target & Llama-3.1-8B-Instruct & NER: mit-movie & Anchor \\
    \midrule
    Source & Qwen2.5-7B-Instruct   &  TweetNER7     & \multirow{5}{*}{Useful Experts} \\
    Source & Qwen2.5-7B-Instruct   &  conll04        & \\
    Source & Qwen2.5-7B-Instruct   &  New York Times & \\
    Source & Qwen2.5-7B-Instruct   &  FabNER         & \\
    Source & Qwen2.5-7B-Instruct   &  FindVehicle    & \\
    \midrule
    Source & Llama-3.1-8B-Instruct &  CrossNER (Science)  & \multirow{4}{*}{Noisy Experts} \\
    Source & Llama-3.1-8B-Instruct &  CrossNER (Politics) & \\
    Source & Llama-3.1-8B-Instruct &  BC4CHEMD (Chemistry)& \\
    Source & Llama-3.1-8B-Instruct &  BC2GM (Biology)     & \\
    \bottomrule
  \end{tabular}
\end{table}

\subsection{GLUE Benchmark}

\begin{table}[H]
  \caption{Performance on the GLUE benchmark.}
  \label{tab:sheet8_glue_summary}
  % \vspace{-2mm}
  \centering
  \setlength{\tabcolsep}{2.0mm}
  \scriptsize
  \begin{tabular}{lcccccc}
    \toprule
    Method & SST-2 & QNLI & RTE & MRPC (F1/Acc) & CoLA & Avg \\
    \midrule
    EMR-Merging & 96.4 & 91.3 & 83.8 & 82.1/75.5 & 44.7 & 79.66 \\
    TIES-Merging & 96.5 & \textbf{91.5} & 83.2 & 81.7/75.2 & 41.6 & 78.90 \\
    Breadcrumbs & 96.5 & 90.6 & 84.7 & 82.0/75.8 & 51.4 & 81.04 \\
    DELLA & \textbf{97.0} & 90.9 & 82.6 & 80.6/73.9 & 38.8 & 77.98 \\
    DARE (TIES+) & 96.7 & 91.0 & 82.7 & 80.4/73.7 & 38.9 & 77.94 \\
    Task Arithmetic & 96.8 & 91.0 & 83.7 & 81.0/74.5 & 45.2 & 79.54 \\
    \midrule
    GAC & 95.2 & 80.1 & 82.3 & 78.4/74.3 & 47.4 & 76.68 \\
    UniTE & 94.8 & 89.1 & 69.6 & 85.2/79.7 & \textbf{61.4} & 80.02 \\
    FuseLLM & 95.2 & 86.8 & 82.3 & 85.5/78.8 & 57.0 & 81.36 \\
    \textbf{HeteroFusion} & 95.3 & 86.4 & \textbf{86.8} & \textbf{85.7}/\textbf{80.3} & 58.7 & \textbf{82.58} \\
    \bottomrule
  \end{tabular}
  \vspace{-1mm}
\end{table}

\section{Mechanistic Insights into HeteroFusion}
\label{sec:mechanistic_insights}

In Section~\ref{sec:methodology}, we introduced the architectural components of \texttt{HeteroFusion}. Here, we provide a deeper mechanistic and theoretical justification for why these specific structural designs successfully mitigate the mismatches inherent in heterogeneous model families.

\subsection{Insights on Topology-based Alignment}
\label{subsec:insights_topology}

\paragraph{Row and Column Structural Views.} 
Directly flattening weight matrices into vectors destroys their algebraic structure as linear operators. Specifically, different LLM families have distinct hidden dimensions, making raw flattened vectors dimensionally incompatible. Inspired by recent advances in weight-space learning \citep{navon2023equivariant}, \texttt{HeteroFusion} avoids this by explicitly projecting the row space (corresponding to the output/projection dimension) and the column space (corresponding to the input/receptive dimension) independently, followed by an additive composition. This design ensures that the extracted representations capture the \emph{local structural statistics} of the functional modules, which are permutation-equivariant and robust to absolute dimensionality changes across different architectures.

\paragraph{Target-Anchored Cross-Attention as an Information Bottleneck.} 
In the HyperNet, we specifically designate the target features as the \emph{Query}, while the concatenated heterogeneous source features serve as the \emph{Key} and \emph{Value}. This asymmetric design is conceptually aligned with the Perceiver architecture \citep{jaegle2021perceiver}. Rather than allowing unconstrained interaction among all heterogeneous experts, the target-as-query mechanism enforces a strict \emph{information bottleneck}. It treats the diverse source models as an unaligned external knowledge dictionary. By computing attention scores based strictly on the target's topological needs, the network inherently filters out architecture-specific noise and selectively retrieves only the task-relevant structural delta that is strictly compatible with the target's latent basis.

\paragraph{Target Adapter Basis Preservation (Updating Only $\bm{B}$).} 
In low-rank adaptation ($\Delta W = BA$), the down-projection matrix $A$ essentially acts as a feature extractor that defines the intrinsic geometric basis of the task's latent subspace, while the up-projection matrix $B$ specifies how these basis vectors are linearly combined for the output \citep{hu2022lora}. In a heterogeneous setting, simultaneously modifying $A_t$ and $B_t$ with cross-architecture signals would permanently alter the target model's native coordinate system, leading to representation collapse and catastrophic interference. By strictly freezing $A_t$ and only predicting structural corrections $\Delta B$, \texttt{HeteroFusion} enforces a strong geometric constraint: all injected heterogeneous knowledge must be expressed as a linear combination of the target's \emph{native} basis vectors. This effectively frames cross-architecture fusion not as an unconstrained structural overwrite, but as a bounded, basis-constrained perturbation (as theoretically shown in Eq.~\ref{eq:stability_bound}), guaranteeing that the core task identity of the target model remains intact.

\subsection{Insights on Conflict-aware Denoising}
\label{subsec:insights_denoising}

\paragraph{SVD-guided Shifted Sparse Gating.} 
Heterogeneous models exhibit severe cross-source conflict because they carry distinct base-model biases encoded during their separate pretraining phases. Existing literature on intrinsic dimensionality \citep{aghajanyan2021intrinsic} demonstrates that task-specific adaptation occurs in a much lower-dimensional subspace than the full parameter space. Our SVD-guided gating mechanism utilizes the singular values of the parameter blocks to estimate their spectral energy distribution. The gating function, equipped with a learnable shift $\mu_{\mathrm{gate}}$, acts as a spectral soft-thresholding operator. It explicitly dampens the spectral channels dominated by architecture-specific variation, preserving only the low-rank invariant structures that are transferable across different model families.

\paragraph{Rectified Distribution Matching (RDM).} 
When features from structurally disjoint models are mapped into a shared latent space, cross-source conflict naturally leads to severe \emph{distribution fragmentation} (i.e., disjoint or severely skewed feature clusters). If left unregularized, this fragmentation causes the subsequent cross-attention to fail due to extreme outliers and incompatible vector geometries. To address this, we introduce RDM regularization, which forces the activated latent representations to match a rectified generalized Gaussian prior. We utilize the Sliced Wasserstein Distance \citep{deshpande2018sliced}, computing mean squared errors over sorted 1D random projections. Similar to the principles of Wasserstein Auto-Encoders \citep{tolstikhin2018wasserstein}, this distribution matching ensures a dense, well-conditioned, and topologically unified manifold, guaranteeing stable cross-source interaction even under high expert disagreement.

\end{document}